\documentclass[10pt,twocolumn,letterpaper]{article}

\usepackage[pagenumbers]{cvpr} 

\newcommand{\TODO}[1]{\textbf{\color{red}[TODO: #1]}}
\renewcommand{\TODO}[1]{}

\usepackage{algorithm}
\usepackage{algorithmic}
\usepackage{xcolor}
\usepackage{xcolor}
\usepackage{multirow}
\usepackage{amsmath, amssymb}
\usepackage[table]{xcolor}
\definecolor{lightgreen}{RGB}{230,255,230}
\definecolor{lightpurple}{RGB}{240, 234, 255}

\floatname{algorithm}{Algorithm} 

\definecolor{cvprblue}{rgb}{0.21,0.49,0.74}
\usepackage[pagebackref,breaklinks,colorlinks,allcolors=cvprblue]{hyperref}

\usepackage[accsupp]{axessibility}  

\def\confName{CVPR}
\def\confYear{2026}

\title{ReConText3D: Replay-based Continual Text-to-3D Generation}

\author{
\normalsize
\begin{tabular}{c@{\hspace{0.8em}}c@{\hspace{0.8em}}c@{\hspace{0.8em}}c}
\href{mailto:muhammad_ahmed_ullah.khan@dfki.de}{\textcolor{black}{Muhammad Ahmed Ullah Khan$^{1,2\dagger}$}} &
\href{mailto:haris.amir@edu.rptu.de}{\textcolor{black}{Muhammad Haris Bin Amir$^{2}$}} &
\href{mailto:didier.stricker@dfki.de}{\textcolor{black}{Didier Stricker$^{1}$}} &
\href{mailto:muhammad_zeshan.afzal@dfki.de}{\textcolor{black}{Muhammad Zeshan Afzal$^{1}$}} \\
\end{tabular} \\ [0.2cm]
\normalsize
\begin{tabular}{c c}
$^{1}$\href{https://dfki.de/web}{\textcolor{black}{DFKI}} &
$^{2}$\href{https://rptu.de/}{\textcolor{black}{RPTU Kaiserslautern-Landau}}
\end{tabular} \\ [0.2cm]
\begin{tabular}{c c}
{\tt\small \href{mailto:muhammad_ahmed_ullah.khan@dfki.de}{\textcolor{black}{muhammad\_ahmed\_ullah.khan@dfki.de}}} & {\tt\small \href{mailto:haris.amir@edu.rptu.de}{\textcolor{black}{haris.amir@edu.rptu.de}}}
\end{tabular}
}

\begin{document}

\twocolumn[{%
\renewcommand\twocolumn[1][]{#1}%
\maketitle
\begin{center}
    \centering
    \captionsetup{type=figure}
    \includegraphics[width=1\linewidth]{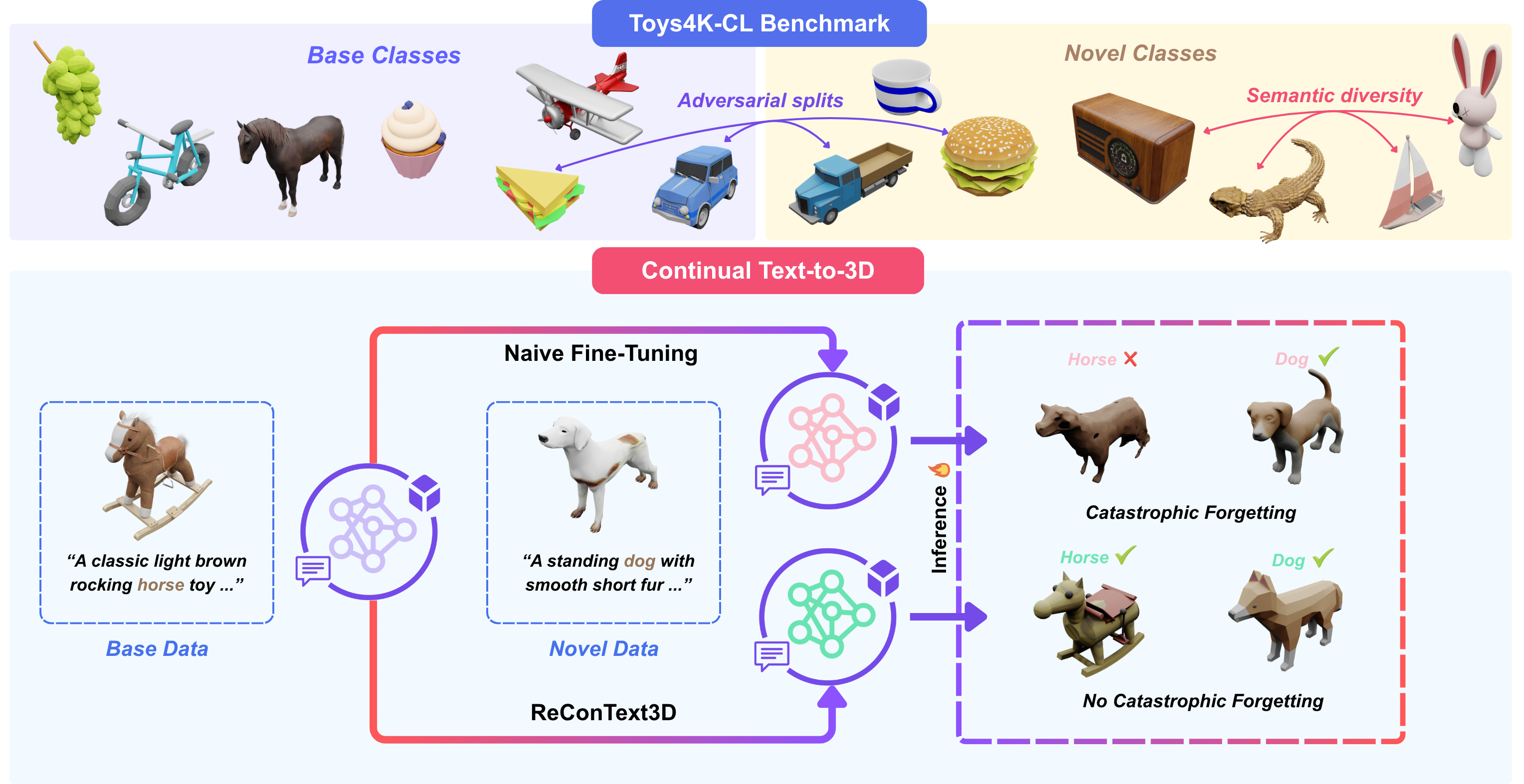}
   \captionof{figure}{\textbf{Top:} The Toys4K-CL benchmark splits data into base classes and novel classes with adversarial splits and semantic diversity while maintaining a comparable number of classes and total assets across both stages. \textbf{Bottom:} Continual text-to-3D generation. Naive fine-tuning on novel data, in this case dog, leads to catastrophic forgetting on base class horse, while continual learning mitigates forgetting, preserving previously learned 3D concepts during model updates.}
   \label{fig:teaser}
   
\end{center}%
}]
 
 \renewcommand{\thefootnote}{\fnsymbol{footnote}}
\footnotetext[2]{\href{mailto:muhammad_ahmed_ullah.khan@dfki.de}{\textcolor{black}{Corresponding Author.}}}
\renewcommand{\thefootnote}{\arabic{footnote}}


\begin{abstract}
Continual learning enables models to acquire new knowledge over time while retaining previously learned capabilities. However, its application to text-to-3D generation remains unexplored. We present \textbf{ReConText3D}, the first framework for continual text-to-3D generation. We first demonstrate that existing text-to-3D models suffer from catastrophic forgetting under incremental training. ReConText3D enables generative models to incrementally learn new 3D categories from textual descriptions while preserving the ability to synthesize previously seen assets. Our method constructs a compact and diverse replay memory through text-embedding k-Center selection, allowing representative rehearsal of prior knowledge without modifying the underlying architecture. To systematically evaluate continual text-to-3D learning, we introduce \textbf{Toys4K-CL}, a benchmark derived from the Toys4K dataset that provides balanced and semantically diverse class-incremental splits. Extensive experiments on the Toys4K-CL benchmark show that \textbf{ReConText3D} consistently outperforms all baselines across different generative backbones, maintaining high-quality generation for both old and new classes. To the best of our knowledge, this work establishes the first continual learning framework and benchmark for text-to-3D generation, opening a new direction for incremental 3D generative modeling. Project page is available at: \href{https://mauk95.github.io/ReConText3D/}{\texttt{https://mauk95.github.io/ReConText3D/}}.
\end{abstract}

\section{Introduction}
\label{sec:intro}

Generative models have made rapid progress in synthesizing 3D assets directly from text, enabling compelling applications in content creation, simulation, and robotics~\cite{jun2023shap,xiang2025structured}. Text-to-3D pipelines based on diffusion or flow objectives can now produce high-quality 3D assets with increasingly faithful text alignment and geometric fidelity~\cite{poole2022dreamfusion,lin2023magic3d,chen2023fantasia3d,wang2023prolificdreamer}. Despite these advances, the dominant training paradigm remains \emph{static}; models are trained once on large, curated datasets and then frozen. In real-world settings, however, a 3D generator must continuously learn new concepts, i.e, categories, shapes, materials, etc., without sacrificing performance on previously learned assets. 


Continual learning (CL) addresses this challenge by training models sequentially over tasks while mitigating \emph{catastrophic forgetting}, the degradation of prior knowledge when learning new data~\cite{mccloskey1989catastrophic,parisi2019continual}. A rich body of work has explored CL for recognition, including rehearsal-based methods that store exemplars~\cite{rebuffi2017icarl}, knowledge distillation~\cite{li2017learning}, and regularization~\cite{li2017learning,zenke2017continual,chaudhry2018riemannian}. In 3D perception, recent efforts such as SDCoT~\cite{zhao2022static} and SDCoT++~\cite{zhao2024sdcot++} extend class-incremental learning to 3D object detection with teacher–student designs and model-agnostic evaluations.

Generative settings pose distinct challenges under continual learning, i.e, distribution shift emerges in both \emph{text} and \emph{shape} spaces, supervision is weakly aligned with perception metrics, and stability–plasticity trade-offs must be managed without sacrificing sample diversity. We argue that continual text-to-3D generation is both practically important and scientifically distinct. Practical pipelines must expand to new product lines or styles over time, while maintaining backward compatibility with existing asset libraries for games, AR/VR, and simulation~\cite{jiang2024survey,li2023generative}. Scientifically, the conditioning pathway (text encoder) and the generative pathway (3D decoder) co-evolve during fine-tuning; naive adaptation on novel classes often drifts the shared conditioning manifold, degrading alignment and appearance for old classes even when the architecture is expressive. 

To the best of our knowledge, \textbf{continual learning for \emph{text-to-3D generation}} has not been studied. In this work, we introduce \textbf{ReConText3D}, the first framework for \emph{continual text-to-3D generation}. ReConText3D is \emph{model-agnostic} and can be plugged into any text-to-3D backbone. The core idea is a simple but effective \emph{semantic replay}: we construct a compact memory of base exemplars using (i) a \emph{count-aware budget allocation} to respect long-tailed data and (ii) \emph{text-embedding k-center selection} to maximize semantic coverage in the prompt space. During novel-stage training, replayed base prompts are mixed with novel data, anchoring the text–shape mapping and mitigating drift, without changing the underlying generative objective or architecture.

To evaluate this new setting, we propose \textbf{Toys4K-CL}, a class-incremental benchmark derived from the captioned Toys4K subset of TRELLIS-500K~\cite{stojanov2021using,xiang2025structured}. We retain 90 well-populated classes and form balanced, semantically diverse base/novel splits, including adversarial arrangements that separate closely related categories to stress interference. We report text alignment, appearance, and geometry metrics following prior work~\cite{xiang2025structured,radford2021learning,qi2017pointnet++}, and analyze forgetting explicitly.

Our \textbf{contributions} can be summarized as follows:
\begin{itemize}[leftmargin=*]
    \item \textbf{Problem \& framework.} We are the first to formulate \emph{continual text-to-3D generation} and present \textbf{ReConText3D}, a simple, model-agnostic framework for continual 3D asset synthesis.
    \item \textbf{Forgetting analysis.} We empirically demonstrate that state-of-the-art text-to-3D models suffer from severe catastrophic forgetting under class-incremental training.
    \item \textbf{Replay method.} We propose a novel replay-based CL method that combines \emph{count-aware budget allocation} with \emph{text-embedding k-center} selection to build compact, semantically diverse memories that mitigate forgetting.
    \item \textbf{Benchmark.} We introduce \textbf{Toys4K-CL}, the first benchmark for \emph{continual text-to-3D generation} with balanced and adversarial base/novel splits .
    \item \textbf{Results.} ReConText3D significantly reduces forgetting and improves overall generation quality across multiple backbones, validating its model-agnostic design.
\end{itemize}
\section{Related Work}
\label{sec:related_work}

\begin{figure*}
  \centering
    \includegraphics[width=0.95\linewidth]{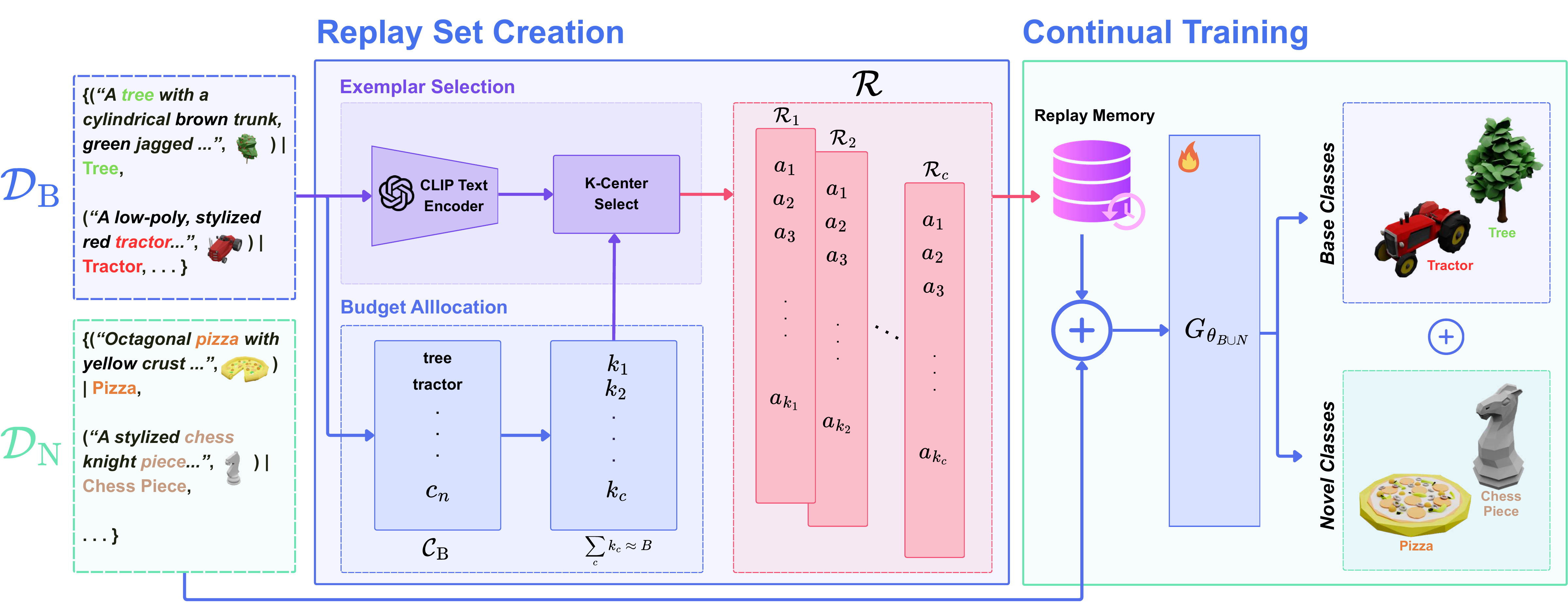}
    \caption{\textbf{Overview of the ReConText3D framework.} Base captions are encoded with CLIP and selected via k-center sampling under a count-aware budget to form replay memory $\mathcal{R}$, which is combined with novel data to incrementally train $G_{\theta_{B\cup N}}$ while preserving base-class synthesis.}
    \label{fig:recontext3d_framework}
     \vspace*{-.6\baselineskip}
\end{figure*}

\textbf{Text-to-3D.}
Recent advances in text-to-3D generation largely fall into two architectural categories, diffusion-based and flow-based generative models. Diffusion-based approaches build upon the success of 2D diffusion models by distilling their learned priors into 3D space. DreamFusion~\cite{poole2022dreamfusion} pioneered this direction using Score Distillation Sampling (SDS) to optimize neural radiance fields under 2D diffusion guidance~\cite{rombach2022high,saharia2022photorealistic} without requiring 3D supervision. Follow-up works~\cite{liang2024luciddreamer, lin2023magic3d,tang2023make,tang2023dreamgaussian, wang2023prolificdreamer} improved geometric consistency and efficiency. Recent models explore native 3D diffusion architectures that operate directly in learned 3D latent spaces. Shap-E~\cite{jun2023shap} introduces a conditional diffusion model that maps text embeddings to implicit 3D representations, enabling rapid text-to-3D asset synthesis after training an encoder on large 3D datasets. 3DTopia-XL~\cite{chen20253dtopia} extends this paradigm by scaling diffusion transformers on a novel primitive-based 3D representation which jointly encodes shape, texture, and material. 

Flow-based approaches~\cite{albergo2022building,lipman2022flow,liu2022flow} have more recently emerged as a powerful class of generative models challenging the dominance of diffusion-based approaches. TRELLIS~\cite{xiang2025structured} employs a flow-transformer architecture trained on large-scale structured 3D latents.

\noindent\textbf{Continual Learning.} Continual learning aims to develop models capable of learning sequentially while mitigating catastrophic forgetting. Major strategies that alleviate forgetting include (1) Replay methods~\cite{castro2018end,hou2019learning,ostapenko2019learning,rebuffi2017icarl,shin2017continual,wu2018memory} that replay old data exemplars during new task training, (2) Regularization-based methods~\cite{chaudhry2018riemannian,dhar2019learning,li2017learning,zenke2017continual} that impose constraints on parameter updates to protect knowledge acquired from previous tasks (3) Structural methods~\cite{mallya2018piggyback,mallya2018packnet,rusu2016progressive} that expand the network architecture or isolate parameters for different tasks to prevent interference. \\
\noindent\textbf{Continual Learning in 3D.} Although continual learning strategies have been explored in the 2D visual domain, including recent advances in text-to-image diffusion models~\cite{smith2023continual,sun2024create,dong2024continually}, their application in 3D is still relatively underexplored. 3D incremental learning has been explored mainly for classification and reconstruction tasks. For classification, methods such as I3DOL~\cite{dong2021i3dol} introduce geometry-aware modules to preserve features across incremental classes, while others employ basic-shape pre-training~\cite{qi2025boosting}, foundation-model adapters~\cite{ahmadi2024foundation}, or spectral exemplar selection~\cite{resani2025continual} to reduce forgetting. For reconstruction, Thai et al.~\cite{thai2022surprising} reveal a positive knowledge transfer effect when sequentially learning 3D shapes from visual inputs. Despite these advances, to the best of our knowledge no prior work addresses continual learning in text-to-3D generation. 

\section{Methodology}
\label{sec:methodology}


\subsection{Problem Definition}
\label{subsection:problem_definition}
In the continual text-to-3D generation setting, we adopt a two-stage (Base→Novel) protocol following standard practice in 3D continual learning \cite{zhao2022static, zhao2022cil}. We consider two non-overlapping sets of classes: a \textbf{base class set} $\mathcal{C}_{\text{base}}$ and a \textbf{novel class set} $\mathcal{C}_{\text{novel}}$, where $\mathcal{C}_{\text{base}} \cap \mathcal{C}_{\text{novel}} = \varnothing$. The corresponding datasets are denoted by $\mathcal{D}_{\text{base}} = \{(t_i, O_i) \mid y_i \in \mathcal{C}_{\text{base}}\}$ and $\mathcal{D}_{\text{novel}} = \{(t_i, O_i) \mid y_i \in \mathcal{C}_{\text{novel}}\}$, where $t_i$ represents a textual description (caption) and $O_i$ is the corresponding 3D asset (mesh).

We define the task of \textbf{continual text-to-3D generation} as: given a text-to-3D generator $G_{\theta_B}$ (the base model) pre-trained on $\mathcal{D}_{\text{base}}$, our goal is to obtain an incremental model $G_{\theta_{B \cup N}}$ by training on $\mathcal{D}_{\text{novel}}$ such that the resulting model can generate 3D assets corresponding to both the base and novel classes, i.e., $\mathcal{C}_{\text{base}} \cup \mathcal{C}_{\text{novel}}$.

In this class-incremental setup, the model must learn new classes from $\mathcal{C}_{\text{novel}}$ while retaining generation capability for previously learned classes in $\mathcal{C}_{\text{base}}$. Following the criteria used in class-incremental learning for classification \cite{rebuffi2017icarl}, during novel training, access to the complete base dataset is not permitted, although limited replay exemplars, i.e, captions may be used depending on the continual learning strategy. This setting simulates a realistic, continual 3D content creation scenario, where models encounter new object categories over time but must avoid catastrophic forgetting of previously learned categories.

\subsection{ReConText3D Framework}
\label{subsection:context3d_framework}

The goal of the proposed \textbf{ReConText3D} framework is to enable \textbf{incremental 3D asset generation from textual descriptions} while mitigating catastrophic forgetting of previously learned classes. As described in \cref{subsection:problem_definition}, given the base dataset $\mathcal{D}_{\text{base}}$ and the novel dataset $\mathcal{D}_{\text{novel}}$, our objective is to train a text-to-3D generator $G_{\theta_{B \cup N}}$ that can generate high-quality 3D meshes for both $\mathcal{C}_{\text{base}}$ and $\mathcal{C}_{\text{novel}}$.

We first train a text-conditioned 3D generative model $G_{\theta_B}$ on the base dataset $\mathcal{D}_{\text{base}}$, learning to generate meshes $O_i \in \mathcal{O}$ conditioned on their textual descriptions $t_i \in \mathcal{T}$ for the base class set $\mathcal{C}_{\text{base}}$. Once $G_{\theta_B}$ converges, we enter the \textbf{continual learning stage}, where the model is incrementally adapted to the novel dataset $\mathcal{D}_{\text{novel}}$ while retaining its ability to generate objects from $\mathcal{C}_{\text{base}}$.

Formally, this adaptation produces an updated generator:
\[
\theta_{B \cup N} \leftarrow \text{CLTrain}\big(G_{\theta_B}, \mathcal{D}_{\text{novel}}, \mathcal{M}_{\text{replay}}\big),
\]
where $\mathcal{M}_{\text{replay}}$ denotes a replay memory containing a subset of caption and 3D mesh pairs from $\mathcal{D}_{\text{base}}$. The combined data from the replay memory and the novel dataset, $\mathcal{M}_{\text{replay}} \cup \mathcal{D}_{\text{novel}}$
, is used to train the incremental model $G_{\theta_{B \cup N}}$, enabling it to revisit representative base-class distributions and preserve generation quality across previously learned concepts.


Our framework is inherently \textbf{model-agnostic} and can be integrated with any text-conditioned 3D generative backbone, independent of the underlying architecture or training objective. It requires only that the model accepts textual prompts as conditioning input and outputs a 3D representation (e.g., mesh, point cloud, or implicit field). Recent text-to-3D backbones either use rectified flow-based methods or latent diffusion-based methods. In our experiments (\cref{sec:experiments}), we demonstrate the versatility of ReConText3D by applying it to one representative model from each type of method, including \textit{TRELLIS-XL}~\cite{xiang2025structured} from rectified flow methods and \textit{Shap-E}~\cite{jun2023shap} from latent diffusion methods. The overall \textbf{ReConText3D} framework, including our proposed replay strategy, is illustrated in \cref{fig:recontext3d_framework}, showing the incremental training process (base~$\rightarrow$~novel).

\subsection{ReConText3D: Replay Set Creation}
\label{subsection:recontext3d}

To mitigate catastrophic forgetting during novel-class adaptation, we propose \textbf{ReConText3D}, a novel replay-based continual generation strategy that selectively reuses representative samples from the base dataset. The core idea is to construct a compact replay memory from the base training set $\mathcal{D}_{\text{base}}$ that preserves semantic diversity (\textbf{text-embedding k-center selection}) while respecting class imbalance (\textbf{count-aware allocation}) inherent to real-world 3D datasets such as Toys4K-CL.

ReConText3D builds the replay memory in two stages: (1) \textbf{budget allocation} and (2) \textbf{exemplar selection}, as illustrated in \cref{alg:allocation} and \cref{alg:kcenter}, respectively. The overall replay set creation process is summarized in \cref{alg:replay_end2end} and illustrated in \cref{fig:recontext3d_framework}.\\

\noindent\textbf{Budget Allocation.} In the allocation stage, we assign replay quotas to each base class based on the square-root of its sample count, using per-class minimum, maximum, and maximum percentage caps $(m_{\min}, m_{\max}, p_{\text{max}})$. This \textbf{$\sqrt{\text{count}}$}-based allocation ensures that frequent classes do not dominate the replay memory while still maintaining adequate representation of underrepresented categories. The resulting per-class allocation $\{k_c\}$ satisfies $\sum_c k_c \approx B$, where $B$ is the global replay budget.\\

\noindent\textbf{Exemplar Selection.} In the selection stage, ReConText3D performs \textbf{k-center selection} in the \textbf{text-embedding space} of asset captions to identify semantically diverse exemplars per class. We use the same text encoder as employed by most text-to-3D backbones, i.e., \textit{CLIP ViT-L/14}~\cite{radford2021learning}, ensuring that the replay memory remains aligned with the textual conditioning space actually perceived by the generator during training. Each asset $a_i$ is represented by the average of its caption embeddings. To allow generality, we denote by $M$ the maximum number of captions considered per asset. We use all available captions ($M=11$) for every asset in the Toys4K-CL benchmark. Given the per-class quota $k_c$, captions are embedded and normalized, and assets are greedily selected to maximize coverage of the text-embedding manifold using cosine similarity in CLIP space.

This design provides a principled balance between \textbf{semantic coverage} and \textbf{class proportionality}. 
Count-aware allocation mitigates long-tail bias by allowing smooth budget growth while preventing large classes from monopolizing memory. 
The k-center strategy ensures diverse semantic coverage so that the replay memory exposes the model to varied textual conditions during continual training.

Considering the constraints of replay-based continual learning, we maintain $B = 248$ samples, i.e replay ratio $r=20\%$ of the novel-set size $N_{\text{novel}}$, with $m_{\min}=3$, $m_{\max}=20$, and $p_{\text{max}}=30\%$. These heuristics were empirically tuned to preserve diversity, avoid overrepresentation of high-frequency classes, and ensure smooth scaling of per-class allocations. 


\begin{algorithm}[t]
\caption{ReConText3D \textsc{CreateReplaySet}}
\label{alg:replay_end2end}
\begin{algorithmic}
\REQUIRE Base metadata with captions $\mathcal{A}$, replay percentage $r$, 
         novel-set size $N_{\text{novel}}$, caps $(m_{\min}, m_{\max}, p_{\max})$
\ENSURE  Replay set $\mathcal{R}$ and associated metadata
\STATE $B \leftarrow \lfloor (r/100)\, N_{\text{novel}} \rfloor$
\STATE $B \leftarrow \min(B, \text{total available})$
\STATE $\{k_c\} \leftarrow$ \textsc{AllocateBudget}($\mathcal{A}, B, m_{\min}, m_{\max}, p_{\max}$)
\STATE $\mathcal{R} \leftarrow \textsc{SelectKCenter}(\mathcal{A}, \{k_c\}, E(\cdot), M)$
\STATE \textbf{return} $\mathcal{R}$
\end{algorithmic}
\end{algorithm}

\begin{algorithm}[t]
\caption{ReConText3D \textsc{AllocateBudget}}
\label{alg:allocation}
\begin{algorithmic}
\REQUIRE Base metadata grouped by class 
         $\mathcal{A}=\{(c,\{a_i\})\}$, global budget $B$, 
         caps $m_{\min}, m_{\max}, p_{\max}\in(0,1]$
\ENSURE  Per-class replay allocation $\{k_c\}$ such that $\sum_c k_c \approx B$

\STATE Initialize $n_c \leftarrow |\{a_i \in c\}|$ \COMMENT{Class availability}

\STATE \textbf{TOTAL}$(\alpha)$:
\STATE \hskip1em $s \leftarrow 0$
\FORALL{class $c$}
    \STATE $u_c \leftarrow \min(m_{\max}, n_c, \lfloor p_{\max} n_c \rfloor)$ \COMMENT{Effective cap}
    \STATE $w \leftarrow \text{clip}(\lfloor \alpha\sqrt{n_c}\rceil, m_{\min}, u_c)$
    \STATE $s \leftarrow s + w$
\ENDFOR
\STATE \hskip1em \textbf{return} $s$

\STATE Find $\alpha$ such that $\text{TOTAL}(\alpha) \approx B$
\STATE \hskip1em Grow or shrink $\alpha$ until $\text{TOTAL}(\alpha) \le B$

\FORALL{class $c$}
    \STATE $k_c \leftarrow \text{clip}(\lfloor \alpha\sqrt{n_c}\rceil, m_{\min}, u_c)$
\ENDFOR

\STATE Greedy rounding adjustment: if $\sum_c k_c > B$ decrement largest $k_c$, else increment smallest
\STATE \textbf{return} $\{k_c\}$
\end{algorithmic}
\end{algorithm}

\begin{algorithm}[t]
\caption{ReConText3D \textsc{SelectKCenter}}
\label{alg:kcenter}
\begin{algorithmic}
\REQUIRE Per-class assets $\{(c,\{(a_i,\mathcal{T}_i)\})\}$, allocations $\{k_c\}$, 
         CLIP text encoder $E(\cdot)$, max captions per asset $M$
\ENSURE  Replay set $\mathcal{R}$ of selected base assets

\STATE \textbf{EmbedAsset}$(a, \mathcal{T})$:
\STATE \hskip1em Select first $M$ valid captions $t_j \in \mathcal{T}$
\STATE \hskip1em $z_j \leftarrow \frac{E(t_j)}{\|E(t_j)\|_2}$, \quad $\bar{z} \leftarrow \frac{1}{M'}\sum_j z_j$
\STATE \hskip1em $v(a) \leftarrow \frac{\bar{z}}{\|\bar{z}\|_2}$ \COMMENT{Average caption embeddings}
\STATE \hskip1em \textbf{return} $v(a)$

\FORALL{class $c$}
    \STATE $V \leftarrow \{v(a_i)\}_{i=1}^{N_c}$, \quad $k \leftarrow k_c$
    \IF{$k \ge N_c$} 
        \STATE $\mathcal{R}_c \leftarrow \{a_i\}$; \textbf{continue}
    \ENDIF
    \STATE $\mu \leftarrow \frac{\sum_i v_i}{\|\sum_i v_i\|_2}$ \COMMENT{Class mean in text space}
    \STATE $s \leftarrow \arg\max_i \langle v_i, \mu \rangle$, \quad $S \leftarrow \{s\}$
    \STATE $d_{\min}(i) \leftarrow 1 - \langle v_i, v_s\rangle$
    \FOR{$t = 2$ to $k$}
        \STATE $u \leftarrow \arg\max_i d_{\min}(i)$
        \STATE $S \leftarrow S \cup \{u\}$
        \STATE $d_{\text{new}}(i)\leftarrow 1-\langle v_i, v_u\rangle$
        \STATE $d_{\min}(i)\leftarrow\min(d_{\min}(i), d_{\text{new}}(i))$
    \ENDFOR
    \STATE $\mathcal{R}_c \leftarrow \{a_i : i\in S\}$
\ENDFOR
\STATE $\mathcal{R} \leftarrow \bigcup_c \mathcal{R}_c$
\STATE \textbf{return} $\mathcal{R}$
\end{algorithmic}
\end{algorithm}

\subsection{Toys4K-CL Benchmark}
\label{subsection:toys4k-cl_benchmark}
To systematically evaluate continual text-to-3D generation, we introduce the \textbf{Toys4K-CL} benchmark, a class-incremental benchmark derived from the Toys4K \cite{stojanov2021using} subset of the TRELLIS-500K dataset~\cite{xiang2025structured}. We select Toys4K as our base due to its clear class annotations, diverse object categories, and its independence from the training data of recent text-to-3D models, making it well-suited for continual evaluation.

\noindent\textbf{Preprocessing and Class Filtering.}
The Toys4K captioned dataset ~\cite{xiang2025structured} contains 3,180 captioned 3D assets (meshes) belonging to 109 natural object classes. To ensure sufficient samples per class for both training and testing, we remove classes containing less than 15 assets, retaining the top 90 classes. Similar to other real-world 3D datasets, our \textbf{Toys4K-CL} dataset exhibits a long-tailed distribution.\\

\noindent\textbf{Base and Novel Class Splits.}
Following the notation introduced in \cref{subsection:problem_definition}, we divide the dataset into disjoint \textit{base} and \textit{novel} class sets, $\mathcal{C}_{\text{base}}$ and $\mathcal{C}_{\text{novel}}$, each containing \textbf{45} classes. To construct these splits, we adopt a balanced and semantically diverse partitioning strategy guided by the following goals:  
(i) maintain a comparable number of classes and total assets across stages;  
(ii) preserve semantic diversity within each stage, ensuring exposure to varied structures, shapes, and materials;  
(iii) provide sufficient training and test samples for each class; and  
(iv) create \textit{adversarial splits} that distribute semantically similar categories (e.g., \textit{dog}, \textit{cat}, \textit{fox}) across the two stages to increase interference and challenge continual learning methods.  

\begin{table*}[t]
\centering
\caption{
\textbf{Quantitative comparision on the Toys4K-CL benchmark.}
We report CLIP similarity (↑) and Fréchet Distance scores computed on Inception and PointNet++ features (↓). 
Forgetting (\%) measures base-class degradation after novel training. Results are reported on TRELLIS-XL and Shap-E backbones. The best results are in bold, and the second-best are underlined.
}

\label{tab:quant_results}
\setlength{\tabcolsep}{5.2pt}
\renewcommand{\arraystretch}{1.05}
\begin{tabular}{c|cccc|cccc|cccc} 
\toprule
\multirow{2}{*}{\textbf{Method}} &
\multicolumn{4}{c|}{\textbf{CLIP} (↑)} &
\multicolumn{4}{c|}{\textbf{FD\textsubscript{Incep}} (↓)} &
\multicolumn{4}{c}{\textbf{FD\textsubscript{Point}} (↓)} \\
\cmidrule(lr){2-5} \cmidrule(lr){6-9} \cmidrule(lr){10-13}
 & Base & Novel & All & F (\%) &
   Base & Novel & All & F (\%) &
   Base & Novel & All & F (\%) \\
\midrule
\multicolumn{13}{c}{\textit{TRELLIS-XL (Flow-based)}} \\
\midrule
Base Training   & 29.60 & –  & –  & - & 73.04 & –  & –  & - & 75.22 & –  & –  & - \\
Fine-tuning     & 24.46 & \underline{29.79} & 27.12 & 17.36 & 101.00 & 75.33 & 56.66 & 38.18 & 72.01 & \underline{69.05} & 70.13 & –4.27 \\
L2-SP           & 24.50 & 29.63 & 27.06 & 17.23 & 102.88 & \underline{75.24} & 57.65 & 40.85 & \textbf{68.47} & 69.05 & \textbf{68.33} & \textbf{–8.98} \\
\rowcolor{lightpurple}
Ours & \underline{28.41} & \textbf{29.86} & \textbf{29.14} & \underline{4.02} & \underline{87.28} & 75.34 & \underline{52.42} & \underline{19.50} & \underline{70.20} & 70.28 & \underline{69.78} & \underline{–6.68} \\
Ours + L2-SP  & \textbf{28.44} & 29.52 & \underline{28.98} & \textbf{3.92} & \textbf{84.85} & \textbf{74.56} & \textbf{51.59} & \textbf{16.17} & 71.57 & \textbf{68.83} & 69.79 & –4.86 \\
Joint Training   & 29.57 & 29.45 & 29.51 & 0.10 & 78.56 & 75.54 & 50.20 & 7.56 & 74.34 & 68.87 & 71.05 & –1.18 \\
\midrule
\multicolumn{13}{c}{\textit{Shap-E (Diffusion-based)}} \\
\midrule
Base Training   & 28.75 & –  & –  & - & 107.11 & –  & –  & - & 25.39 & –  & –  & - \\
Fine-tuning     & 27.80 & \textbf{28.70} & 28.25 & 3.30 & 118.28 & 108.99 & 85.43 & 10.43 & 25.46 & \underline{22.05} & 23.37 & 0.28 \\
L2-SP           & 27.38 & 28.27 & 27.83 & 4.77 & 123.98 & 112.87 & 89.92 & 15.75 & 25.76 & 22.25 & 23.58 & 1.49 \\
\rowcolor{lightpurple}
Ours & \textbf{28.34} & \underline{28.54} & \textbf{28.44} & \textbf{1.43} & \textbf{110.49} & 111.77 & \textbf{83.26} & \textbf{3.15} & \textbf{25.21} & \textbf{21.78} & \textbf{23.17} & \textbf{–0.67} \\
Ours + L2-SP  & \underline{28.33} & 28.47 & \underline{28.40} & \underline{1.46} & \underline{112.31} & \underline{111.00} & \underline{83.84} & \underline{4.85} & \underline{25.25} & 22.35 & \underline{23.45} & \underline{–0.55} \\
Joint Training   & 28.45 & 28.53 & 28.49 & 1.04 & 112.23 & 110.16 & 83.33 & 4.77 & 25.65 & 22.28 & 23.63 & 1.05 \\
\bottomrule
\end{tabular}
\vspace*{-.6\baselineskip}
\end{table*}

To design these splits, we prompted GPT4o~\cite{hurst2024gpt} to propose balanced partitions from the full class distribution under the above constraints. This results in two 45-class splits, Base and Novel, with near-uniform class counts and complementary semantic coverage. Each class contributes up to five assets for testing, with the remaining samples used for training. Our test split consists of \textbf{450} samples, with \textbf{225} samples for both base and novel classes. The training set has \textbf{1352} samples for base class assets and \textbf{1243} samples for novel class assets. 
\newline\newline\textbf{Benchmark Properties.}
Toys4K-CL provides a compact yet challenging continual generation benchmark featuring realistic long-tailed data statistics, class imbalance, and inter-class similarity—factors that collectively stress-test catastrophic forgetting and knowledge transfer in generative models. Detailed class lists and per-split statistics are provided in the Supplementary \cref{section:supp_toys4k-cl_benchmark}.

\section{Experiments}
\label{sec:experiments}

\subsection{Implementation Details}
\label{subsection:implementation_details}
\noindent\textbf{Backbones.}
We experiment with two representative text-to-3D generation backbones: \textit{TRELLIS-XL} (flow-based model) and \textit{Shap-E} (diffusion-based model), enabling analysis across archi-
tectures rather than model variants. For both models, we only train their text-conditioned parts of the models. For TRELLIS, we train its text-conditioned generative models, the Sparse-Structure (SS) Flow model and Structured-Latent (SLAT) Flow model. The SS and SLAT representations required for training are generated using the pretrained VAEs. We also use the pretrained mesh decoder to decode the generated SLATs into 3D meshes. Both VAEs and mesh decoder are already pre-trained on the TRELLIS-500K train dataset (Toys4K not included) and open-sourced~\cite{xiang2025structured}. For Shap-E, we use the pretrained SDF-VAE, provided by~\cite{luo2023scalable} to obtain latent volumes and train only its text-conditioned diffusion model.

\noindent\textbf{Training setup.} For TRELLIS-XL, the base model is trained from scratch for 360k and 150k steps for the SS and SLAT models, respectively, with a learning rate of $1\mathrm{e}{-4}$. Continual-learning baselines are fine-tuned for 200k (SS) and 120k (SLAT) steps with a learning rate of $1\mathrm{e}{-5}$. All the models are trained on 4 A100-80GB GPUs with a batch size of 8 and 24 for the SS and SLAT flow model, respectively.
For Shap-E, the text-conditioned diffusion model is trained using the weights from a pre-trained checkpoint provided by ~\cite{luo2023scalable}. 
Both the base model training and the continual learning baselines are trained for 1000 epochs with a learning rate of $1\mathrm{e}{-5}$, and the best checkpoint is used for evaluation. All models are trained on a single H100 GPU with a batch size of 16. All other training hyperparameters are the same as mentioned in the original papers for both TRELLIS~\cite{xiang2025structured} and Shap-E~\cite{jun2023shap}. All the models were trained on our \textbf{Toys4K-CL} training set. Originally, each asset has 11 captions (different levels of details). Similar to TRELLIS, we randomly select 1 out of 11 during training. For evaluation, the most descriptive caption is used.

\noindent\textbf{Remarks.}
Unlike TRELLIS, we did not train the diffusion-based Shap-E generative model from scratch, as it failed to converge on the relatively small Toys4K-CL dataset. The flow-based TRELLIS generators use simpler objectives and two-stage generation (sparse→structured), whereas Shap-E’s one-stage diffusion objective is considerably more complex, motivating the use of pretrained checkpoints.

\subsection{Baselines}
\label{subsection:baselines}
We compare \textbf{ReConText3D} against the following continual-learning and reference baselines:

\begin{itemize}[leftmargin=*]
    \item \textbf{Base Training:} model trained only on $\mathcal{D}_{\text{base}}$; serves as the initial representation and upper-bound performance for base classes.
    \item \textbf{Fine-tuning:} direct fine-tuning of the base model on $\mathcal{D}_{\text{novel}}$ without any constraint. Updates all parameters of the base model.
    \item \textbf{L2-SP}~\cite{xuhong2018explicit}: fine-tuning with an additional L2-SP regularization term applied to all learnable parameters except bias and normalization layers. Serves as the comparision with a known continual learning method.
    \item \textbf{ReConText3D (ours) + L2-SP:} combination of L2-SP with our ReConText3D replay strategy.
    \item \textbf{Joint Training:} model trained from scratch on $\mathcal{D}_{\text{base}}\!\cup\!\mathcal{D}_{\text{novel}}$; represents the upper performance bound.
\end{itemize}

\begin{figure*}
  \centering
    \includegraphics[width=0.95\linewidth]{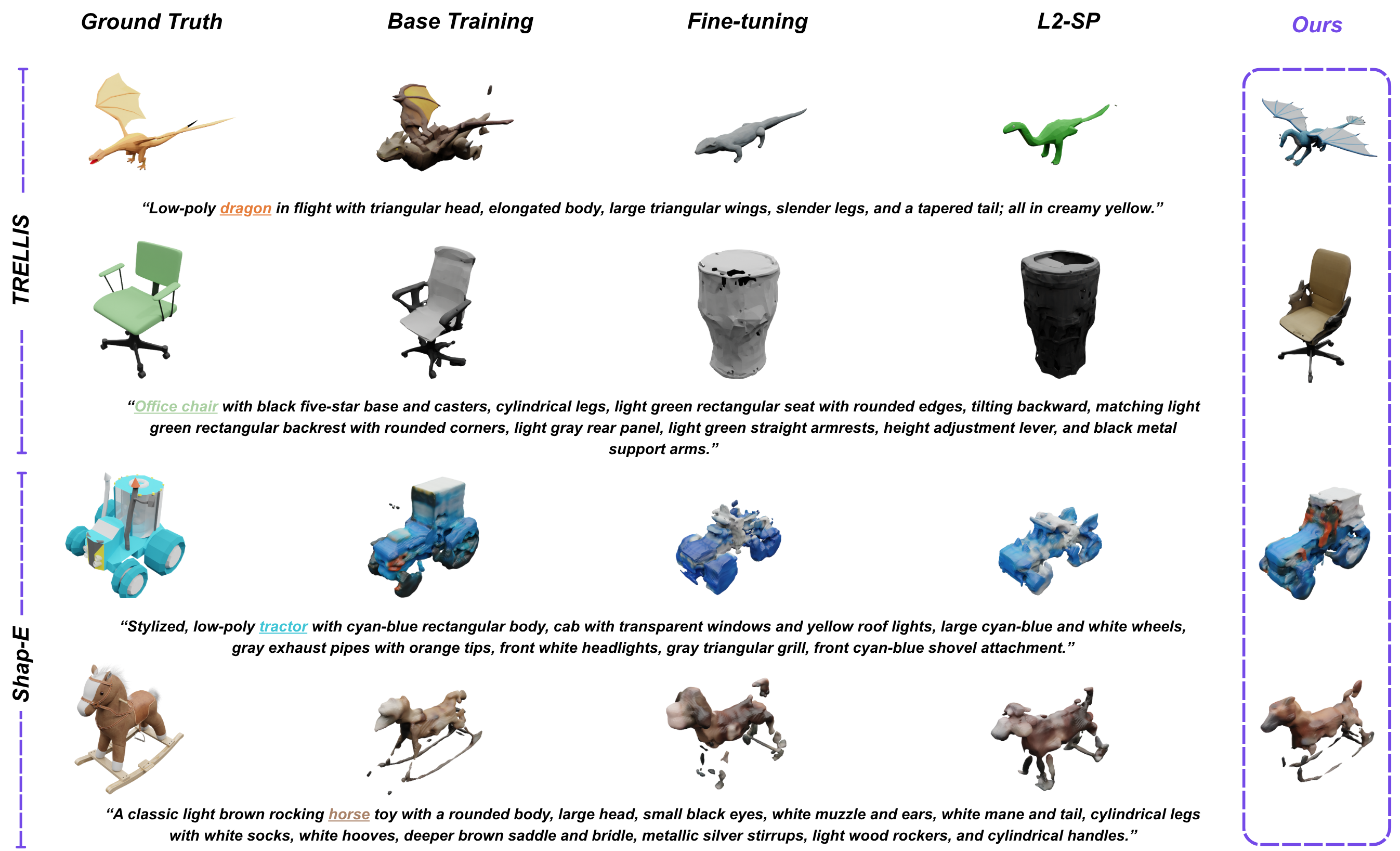}
    \caption{\textbf{Qualitative comparision of continual text-to-3D baselines on base class assets:} From left to right: Ground Truth, Base Training, Fine-tuning, L2-SP, and our method. Results on TRELLIS are shown on the top, while Shap-E is shown on the bottom.}
    \label{fig:qual_results_base_class}
    \vspace*{-.6\baselineskip}
\end{figure*}

\subsection{Quantitative Results}
\label{subsection:quantitative_results}


\noindent\textbf{Evaluation Metrics.}
For quantitative evaluation of the generated 3D assets, we use the metrics reported by TRELLIS~\cite{xiang2025structured}. We report CLIP score~\cite{radford2021learning} to measure text-mesh alignment and Frechet Distance (FD), using Inception-v3~\cite{szegedy2016rethinking} and PointNet++~\cite{qi2017pointnet++} feature embeddings, for appearance and geometric quality, respectively. Forgetting (F) is measured as the relative percentage drop in base-class performance after novel training.


\noindent\textbf{Results on TRELLIS-XL and Shap-E.} Table~\ref{tab:quant_results} presents the quantitative comparision on the test set of our \textbf{Toys4K-CL} benchmark for both TRELLIS-XL and Shap-E models. Naive fine-tuning exhibits clear \textit{catastrophic forgetting}, with strong performance degradation on previously learned (Base) classes, i.e., a \textbf{17.4\%} drop in CLIP score and \textbf{38.2\%} forgetting in $\text{FD}_\text{Incep}$ on TRELLIS-XL, and smaller but consistent losses of \textbf{3.3\%} in CLIP score and \textbf{10.4\%} in $\text{FD}_\text{Incep}$ on Shap-E. 
In contrast, our proposed \textbf{ReConText3D} replay strategy markedly reduces forgetting and improves overall generative quality. 
On TRELLIS-XL, ReConText3D reduces the CLIP forgetting by $\approx$\textbf{77\%} and $\text{FD}_\text{Incep}$ forgetting by $\approx$\textbf{50\%} compared to naive fine-tuning. Similarly, on Shap-E, it lowers the forgetting by $\approx$\textbf{56\%} in CLIP score and $\approx$\textbf{70\%} in $\text{FD}_\text{Incep}$ forgetting compared to the fine-tuning baseline, approaching the joint-training upper bound.

Beyond mitigating forgetting, \textbf{ReConText3D} consistently enhances the overall performance across the base and novel classes, indicating that exemplar-based replay not only preserves past knowledge but also facilitates better generalization to new classes. We attribute this to a \textit{semantic rehearsal effect}; by retaining a compact and diverse set of representative Base samples in text-embedding space, the model maintains alignment between the learned text-conditioned priors and its generative latent distribution, preventing drift in the shared conditioning space. This replayed supervision anchors the text-to-3D mapping and stabilizes training dynamics, especially under continual fine-tuning on limited novel-class data.

When combined with a mild \textbf{L2-SP} regularization term, the hybrid \textbf{ReConText3D + L2-SP} variant further reduces the forgetting, achieving the best scores for base classes on CLIP and $\text{FD}_\text{Incep}$ metric for TRELLIS-XL. 

Interestingly, we observe \textit{negative forgetting} (i.e., positive backward transfer) on the geometric $\text{FD}_\text{Point}$ metric across both models, suggesting that learning novel structural categories helps refine global 3D shape priors, improving geometry fidelity even for Base classes. Class-wise results are presented in the Supplementary \cref{section:supp_quantitative_results}.


\subsection{Qualitative Results}
\label{subsection:qualitative_results} 
\begin{figure}[t]
  \centering
    \includegraphics[width=1.0\linewidth]{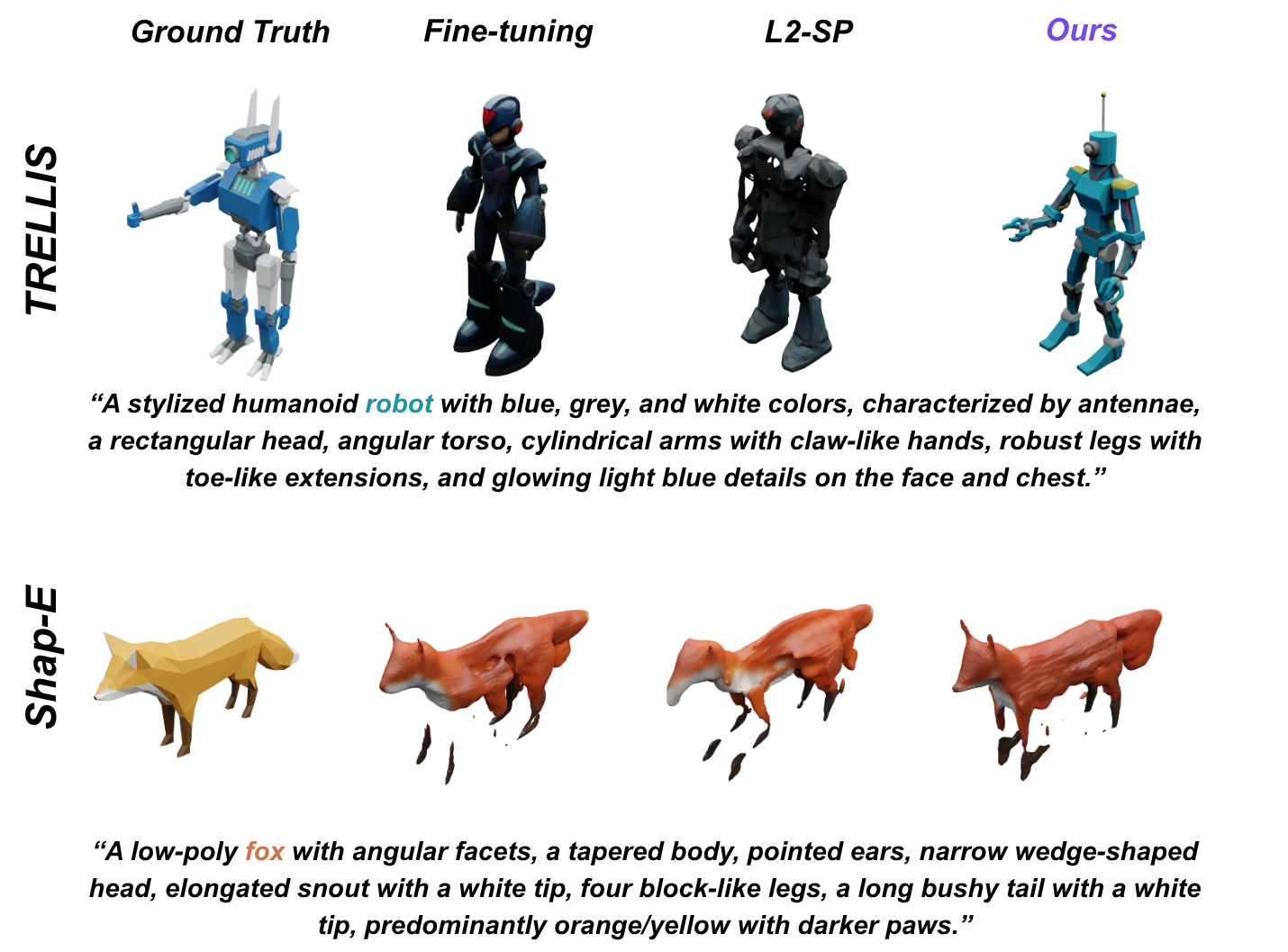}
   \caption{\textbf{Qualitative comparision} of novel class assets across baselines. }
   \label{fig:qual_results_novel_class}
   \vspace*{-.6\baselineskip}
\end{figure}

\cref{fig:qual_results_base_class} and \cref{fig:qual_results_novel_class} present qualitative comparisions of meshes generated from representative text prompts of the base and novel classes, respectively. Fine-tuning exhibits visible degradation on base classes, while L2-SP partially preserves geometry but fails to maintain texture fidelity. Fine-tuning fails to keep the semantic understanding of the base classes and confuses them with novel classes. In the first example (row 1), it generates a lizard, which is a novel class, instead of the original dragon class from the base set. ReConText3D, on the contrary, retains semantic understanding and structural detail for old classes. Additionally, as seen in \cref{fig:qual_results_novel_class}, ReConText3D successfully synthesizes plausible novel objects, with better quality compared to fine-tuning baseline, illustrating an effective balance between stability and plasticity in 3D generation. Extensive qualitative results are presented in the Supplementary \cref{section:supp_qualitative_results}.

\subsection{Ablation Studies}
\label{subsection:ablations}
\begin{figure}[t]
    \centering
    \includegraphics[width=\linewidth]{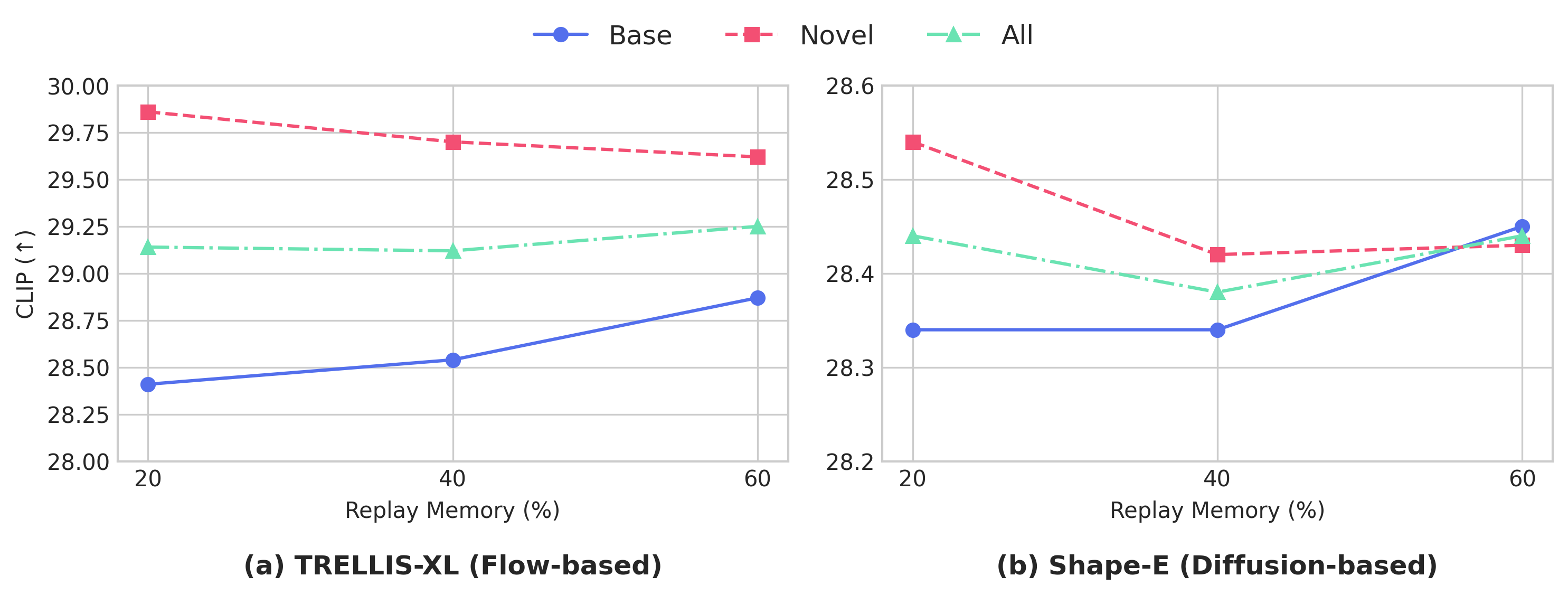}
    \caption{
    \textbf{Ablation on replay memory size.}
    Performance of our replay strategy under varying replay memory budgets/sizes
    on TRELLIS-XL and Shap-E backbones.
    }
    \label{fig:ablations_replay_size}
    \vspace*{-.6\baselineskip}
\end{figure}

\begin{figure}[t]
    \centering
    \includegraphics[width=\linewidth]{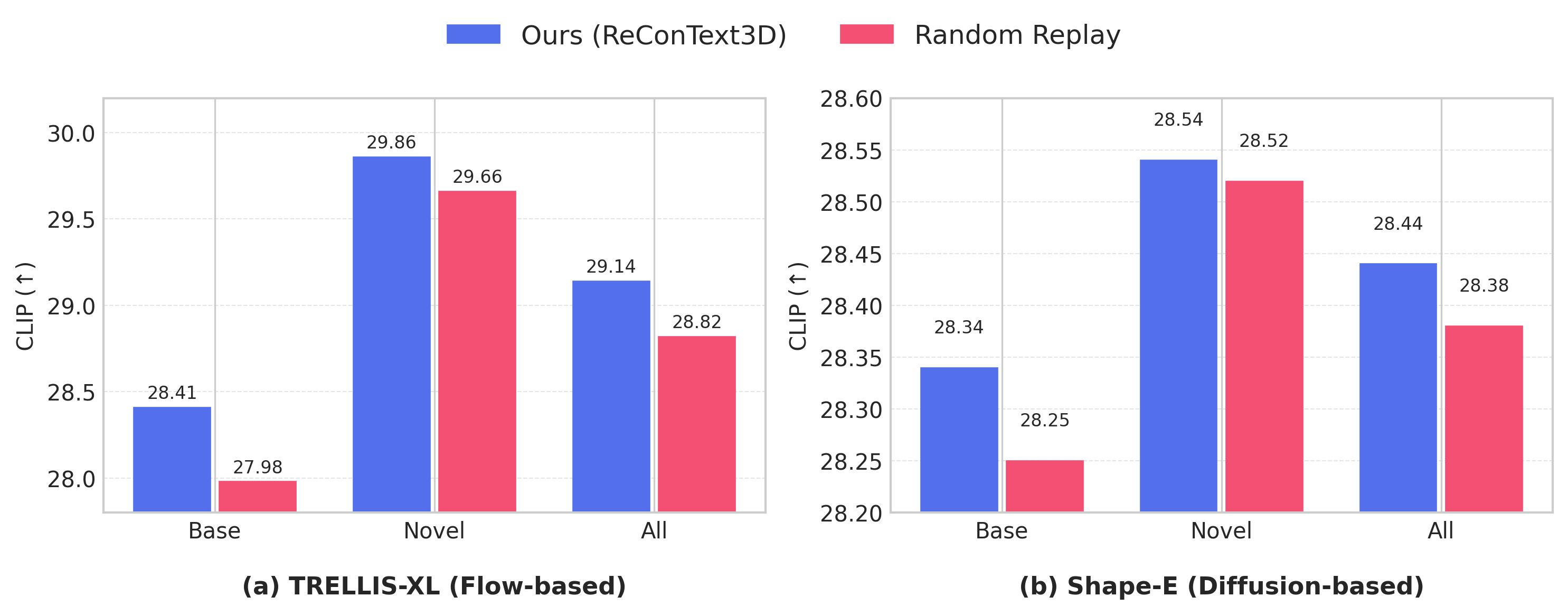}
    \caption{
    \textbf{Ablation on replay strategy.}
    comparision with random replay strategy on TRELLIS-XL and Shap-E backbones.
    }
    \label{fig:ablations_replay_type}
    \vspace*{-.6\baselineskip}
\end{figure}

We present ablation studies of our ReConText3D method in Fig~\ref{fig:ablations_replay_size} and Fig~\ref{fig:ablations_replay_type} to investigate the effects of: size of replay memory and type of replay strategy.\\

\noindent\textbf{Replay Memory Size.} We analyze the influence of the replay memory budget $B$ in our method using TRELLIS-XL and Shap-E model. For this, we create three replay sets with  $r\!\in\!\{20\%,40\%,60\%\}$ of the novel-set size $N_{novel}$. As shown in ~\cref{fig:ablations_replay_size}, CLIP performance on the base set improves steadily with larger replay buffers. Suggesting that a compact memory is sufficient for stable, continual generation.

\noindent\textbf{Replay Type.} To evaluate our k-Center selection, we compare it against a simple replay-based strategy, \textit{Random Replay}, where replay samples are randomly selected from the entire base dataset. Results in ~\cref{fig:ablations_replay_type} show that our text-embedding k-Center replay achieves the best overall balance between base retention and novel adaptation in terms of CLIP scores, demonstrating the importance of semantically diverse and representative replay construction in text-conditioned 3D generation.

\section{Limitations}
\label{sec:limitations}

While \textbf{ReConText3D} shows strong performance and cross-architecture generalizability, our study is limited to a two-stage class-incremental setup with a fixed replay budget. Extending to multi-stage continual learning would require larger 3D datasets to maintain sufficient samples per class, particularly crucial for diffusion-based models like Shap-E, which demand substantial data for stable convergence. Broader object domains may also introduce challenges such as replay-memory saturation and semantic drift over longer training horizons. Moreover, current text-to-3D evaluations still rely on 2D-adapted metrics, future work should therefore explore perceptual and physical realism metrics defined directly in 3D space for more faithful assessment.

\section{Conclusion}
\label{sec:conclusion}

This work, for the first time, studies catastrophic forgetting in existing text-to-3D models under continual learning. We presented \textbf{ReConText3D}, the first framework for continual text-to-3D generation, enabling generative models to incrementally learn new 3D categories while retaining prior knowledge. Our replay-based strategy combines count-aware allocation with text-embedding k-center selection, achieving a strong balance between stability and plasticity without altering the base architecture. To support systematic evaluation, we introduced the \textbf{Toys4K-CL} benchmark with balanced and semantically diverse base–novel splits. Experiments across multiple backbones demonstrate that ReConText3D significantly mitigates catastrophic forgetting, confirming its model-agnostic effectiveness. We hope this work lays a foundation for future research on continual 3D generation.


\section{Acknowledgment}
\label{sec:acknowledgment}

This work was co-funded by the European Union under Horizon Europe, grant number 101135724, project LUMINOUS. However, the views and opinions expressed are those of the author(s) only and do not necessarily reflect those of the European Union. Neither the European Union nor the granting authority can be held responsible.
{
    \small
    \bibliographystyle{ieeenat_fullname}
    \bibliography{main}
}

\clearpage
\setcounter{page}{1}
\maketitlesupplementary


This supplementary document provides additional details supporting the main paper. We first present further information on the design and statistics of the \textbf{Toys4K-CL} benchmark (\cref{section:supp_toys4k-cl_benchmark}), including semantic grouping and long-tailed class distributions. We then describe our \textbf{ReConText3D replay construction} and illustrate how budget allocation affects class coverage (\cref{section:supp_replay_creation}). Next, we report full \textbf{class-wise quantitative results} for both TRELLIS-XL and Shap-E across base and novel categories (\cref{section:supp_quantitative_results}). Finally, we include extended \textbf{qualitative comparisons} and highlight representative failure cases (\cref{section:supp_qualitative_results}). Together, these materials complement the main paper with detailed analysis and visualizations.

\section{Additional Details on Toys4K-CL Benchmark}
\label{section:supp_toys4k-cl_benchmark}
In this section, we present the additional details regarding our \textbf{Toys4K-CL Benchmark}.\\

\begin{figure*}[t]
    \centering
    \begin{subfigure}{0.48\linewidth}
        \centering
        \includegraphics[width=\linewidth]{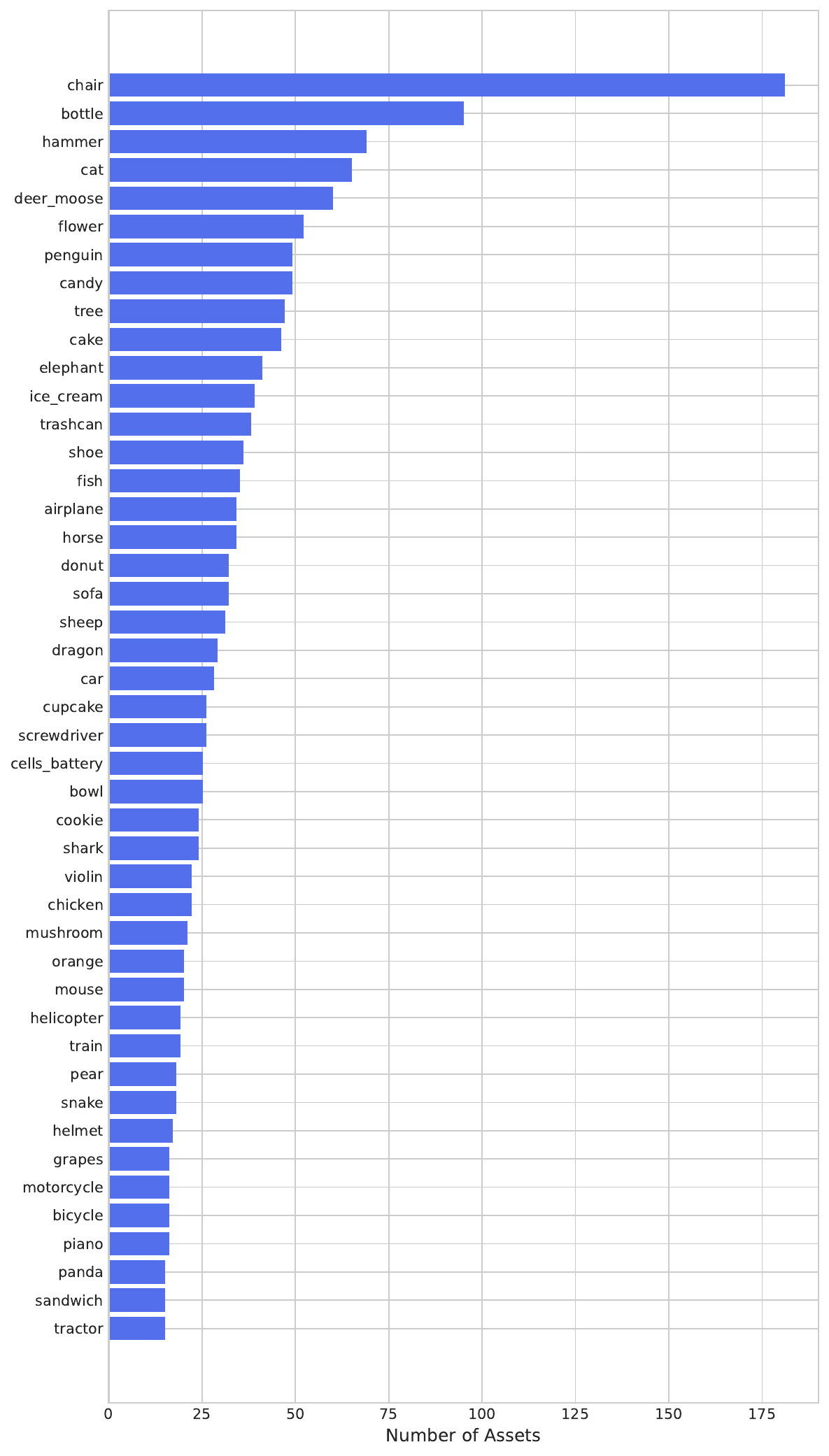}
        \caption{Base classes}
        \label{fig:stage1_dist}
    \end{subfigure}
    \hfill
    \begin{subfigure}{0.48\linewidth}
        \centering
        \includegraphics[width=\linewidth]{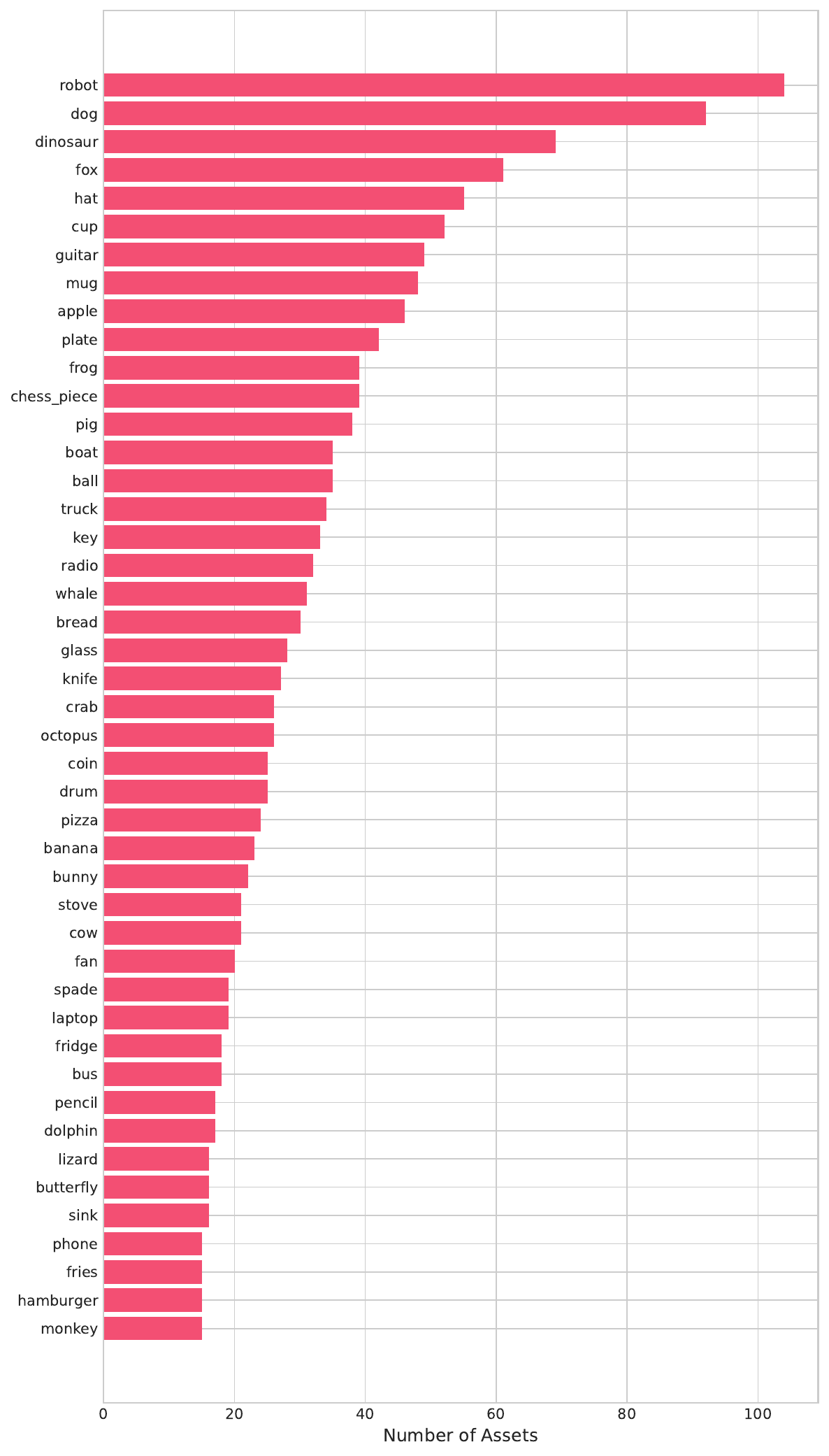}
        \caption{Novel classes}
        \label{fig:stage2_dist}
    \end{subfigure}
    \caption{\textbf{Class distribution statistics for the Toys4K-CL benchmark (train and test).} 
    Both splits exhibit similar long-tailed behaviour.}
    \label{fig:combined_distribution}
\end{figure*}

\noindent\textbf{Benchmark details.}
Toys4K-CL consists of 45 base and 45 novel classes selected to maximize semantic diversity while preserving the natural long-tailed distribution of the Toys4K dataset \cite{stojanov2021using}. Based on an analysis of the dataset’s taxonomy, we observed that assets naturally cluster into the following broad semantic groups: \textbf{furniture/households} (\emph{chair}, \emph{sofa}, \emph{fridge}), \textbf{tools/utensils} (\emph{hammer}, \emph{screwdriver}, \emph{pencil}), \textbf{vehicles} (\emph{airplane}, \emph{car}, \emph{truck}), \textbf{animals/creatures} (\emph{dog}, \emph{fox}, \emph{dragon}), \textbf{food} (\emph{banana}, \emph{cookie}, \emph{pizza}), and \textbf{instruments/others} (\emph{guitar}, \emph{piano}, \emph{monitor}), and constructed splits such that each contains categories spanning all groups. This ensures broad variation in geometry, appearance, and topology across both base and novel classes. \cref{tab:benchmark_stats_suppl} reports per-split statistics, including the counts of classes, train, test, and total samples. \cref{fig:combined_distribution} visualizes the class-frequency distributions for base and novel splits. Both splits exhibit similar long-tailed behaviour, reflecting realistic class imbalance conditions.

\begin{table}[h]
\centering
\small
\caption{\textbf{Toys4K-CL Benchmark Statistics.}}
\label{tab:benchmark_stats_suppl}
\setlength{\tabcolsep}{10pt}
\renewcommand{\arraystretch}{1.1}
\begin{tabular}{lcccc}
\toprule
\textbf{Split} & \textbf{Classes} & \textbf{Train} & \textbf{Test} & \textbf{Total} \\
\midrule
Base  & 45 & 1352 & 225 & 1577 \\
Novel & 45 & 1243 & 225 & 1468 \\
\bottomrule
\end{tabular}
\end{table}

\section{Additional Details on ReConText3D Replay Creation}
\label{section:supp_replay_creation}

\begin{figure}[t]
  \centering
    \includegraphics[width=1.0\linewidth]{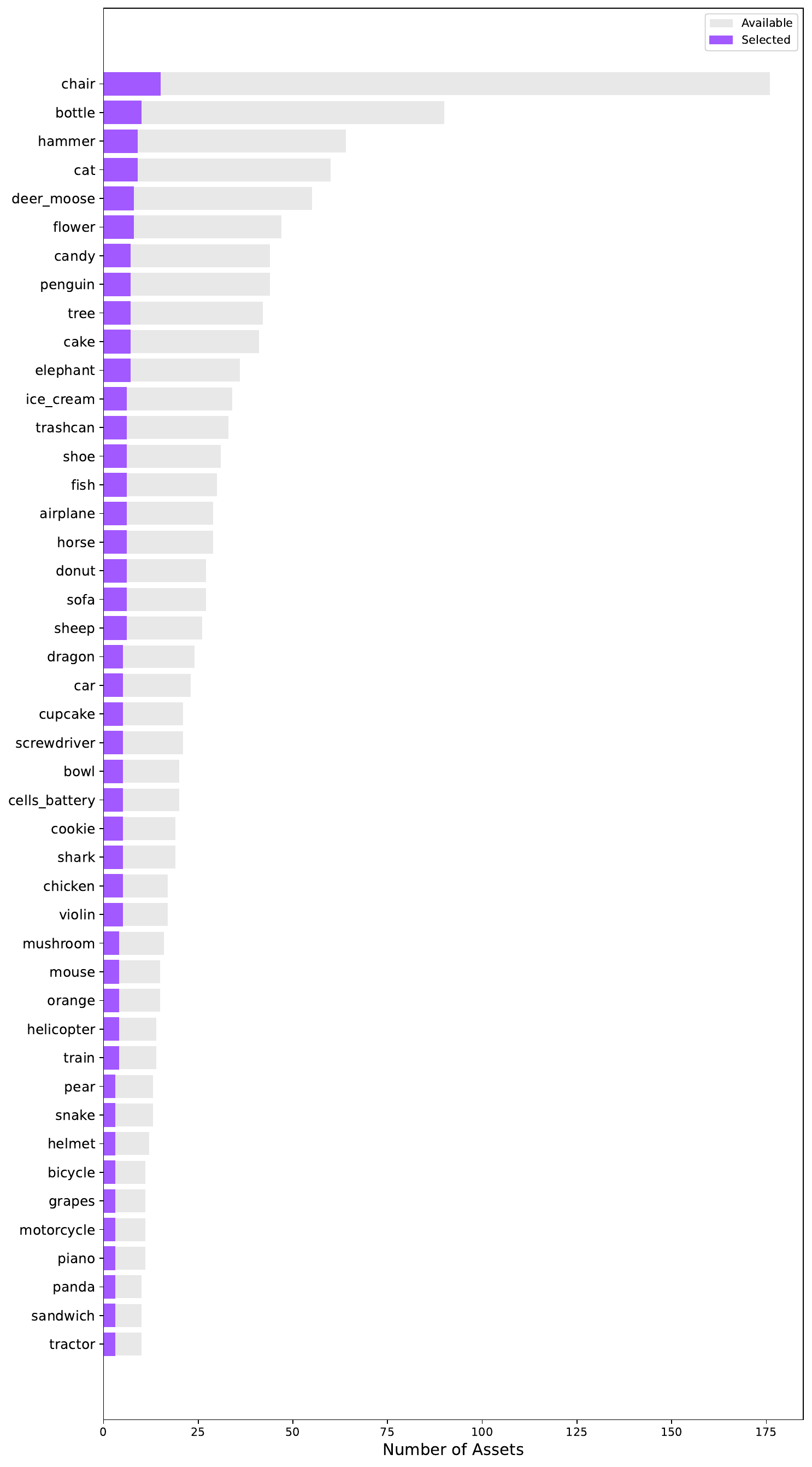}
   \caption{Number of base class assets (replay examplers) returned by our count-aware budget allocation}
   \label{fig:toys4k_replay_selection_distribution_horizontal}

\end{figure}

Our ReConText3D replay strategy provides a principled balance between semantic coverage and class proportionality of the replayed examplers. Figure~\ref{fig:toys4k_replay_selection_distribution_horizontal} shows that our count-aware allocation mitigates long-tail bias by allowing smooth budget growth while preventing large classes from monopolizing memory.

\section{Additional Quantitative Results}
\label{section:supp_quantitative_results}

In this section, we provide detailed class-wise CLIP scores for both base and novel categories on the Toys4K-CL benchmark, complementing the aggregated results in the main paper.

\paragraph{Base-class performance.}
Table~\ref{tab:quant_classwise_base_clip} reports class-wise CLIP \cite{radford2021learning} scores on base-class assets for both \textbf{TRELLIS-XL} \cite{xiang2025structured} (left) and \textbf{Shap-E} \cite{jun2023shap} (right) across the baselines. For each backbone, the \textit{Ours} column (ReConText3D replay) is highlighted in light purple. Overall, ReConText3D either matches or improves over naive fine-tuning and L2-SP \cite{xuhong2018explicit} on the vast majority of classes, while remaining competitive with or close to the base-training upper bound. On TRELLIS-XL, this trend is particularly visible for categories that are prone to strong forgetting, such as \emph{airplane}, \emph{chair} and \emph{grapes}, where fine-tuning substantially degrades CLIP similarity, whereas ReConText3D restores or surpasses the original base-training scores. On Shap-E, ReConText3D similarly offers consistent gains or stability across diverse object types (e.g. \emph{hammer}, \emph{panda}, \emph{pear}), confirming that text-space replay remains effective even for diffusion-based generators. The joint behavior of L2-SP+Ours also illustrates the complementary effect of weight regularization and semantic replay: in some classes L2-SP+Ours can be slightly higher, whereas in many others, ReConText3D alone achieves the best performance.

\paragraph{Novel-class performance.}
Table~\ref{tab:quant_classwise_novel_clip} presents the corresponding CLIP scores for novel-class assets. Here again, we compare fine-tuning, L2-SP, L2-SP+Ours, and our replay strategy for both TRELLIS-XL and Shap-E, with the ReConText3D columns highlighted. While the primary goal of replay is to preserve base-class performance, the results show that ReConText3D also maintains strong or even improved performance on novel classes. For many categories (e.g., \emph{apple}, \emph{fries}, \emph{pencil}, \emph{whale}), the ReConText3D configuration either achieves the best CLIP score or remains very close to the fine-tuning baseline, indicating that replay does not overly constrain plasticity. This pattern is consistent across both backbones and across both simple and visually complex categories (such as \emph{chess\_piece}, \emph{glass}, \emph{radio}), supporting our claim that semantic replay can balance stability and plasticity without sacrificing generative alignment for novel concepts.

Overall, the class-wise results on base and novel assets demonstrate that ReConText3D yields robust improvements across a wide range of object categories and model families. Rather than benefiting only a small subset of classes, the replay mechanism consistently stabilizes base performance while preserving (and often enhancing) novel-class quality, providing fine-grained evidence for the global trends reported in the main paper.

\paragraph{Failure cases and challenging categories.}
While ReConText3D delivers consistent improvements across most classes, the class-wise results also reveal a small number of categories where replay remains challenging. These cases typically arise in highly fine-grained or visually ambiguous object types, or in categories with inherently low intra-class consistency. For example, on TRELLIS-XL, classes such as \emph{dragon}, \emph{mouse}, and \emph{tractor} show only modest gains or slight reductions compared to the strongest baseline, suggesting that their visual and geometric variability may reduce the benefit of text-embedding replay. Similarly, on Shap-E, categories such as \emph{helicopter}, \emph{cat}, and \emph{orange} exhibit limited improvements.

\begin{table*}[t]
\centering
\small
\caption{\textbf{Class-wise Evaluation (CLIP (↑))} on Base-Class assets for TRELLIS-XL and Shap-E. Best scores are highlighted in \textbf{bold}, whereas our ReConText3D results are highlighted in purple. Best scores are shown excluding Base Training.}
\label{tab:quant_classwise_base_clip}
\setlength{\tabcolsep}{3pt}
\renewcommand{\arraystretch}{1.1}
\begin{tabular*}{\textwidth}{@{\extracolsep{\fill}}l ccccc|ccccc}
\toprule
\multirow{2}{*}{\textbf{Class}} &
\multicolumn{5}{c}{\textit{TRELLIS-XL}} &
\multicolumn{5}{c}{\textit{Shap-E}} \\
\cmidrule(lr){2-6} \cmidrule(lr){7-11}
& Base Training & Fine-tuning & L2-SP & L2SP + Ours & \cellcolor{lightpurple}\textbf{Ours} 
& Base Training & Fine-tuning & L2-SP & L2SP + Ours & \cellcolor{lightpurple}\textbf{Ours} \\
\midrule

airplane        & 29.37 & 22.83 & 21.52 & \textbf{30.85} & \cellcolor{lightpurple}30.77
                & 30.83 & 25.14 & 23.09 & \textbf{30.54} & \cellcolor{lightpurple}29.76 \\
bicycle         & 27.21 & 18.12 & 19.04 & \textbf{25.88} & \cellcolor{lightpurple}25.36
                & 27.18 & 26.17 & 23.33 & 25.40 & \cellcolor{lightpurple}\textbf{26.19} \\
bottle          & 32.96 & 28.54 & 24.38 & \textbf{30.66} & \cellcolor{lightpurple}29.75
                & 31.52 & \textbf{31.37} & 31.30 & 30.53 & \cellcolor{lightpurple}31.18 \\
bowl            & 33.50 & 30.97 & 31.38 & \textbf{33.94} & \cellcolor{lightpurple}33.78
                & 33.86 & 32.81 & 32.81 & 33.68 & \cellcolor{lightpurple}\textbf{33.85} \\
cake            & 28.12 & 21.60 & 21.68 & 27.06 & \cellcolor{lightpurple}\textbf{27.35}
                & 29.28 & 28.63 & 28.70 & \textbf{29.57} & \cellcolor{lightpurple}29.49 \\
candy           & 34.72 & 26.95 & 28.99 & \textbf{29.06} & \cellcolor{lightpurple}28.33
                & 34.64 & 32.20 & 32.60 & \textbf{34.08} & \cellcolor{lightpurple}33.88 \\
car             & 26.87 & \textbf{28.36} & 28.13 & 27.71 & \cellcolor{lightpurple}28.35
                & 28.32 & 27.29 & 27.01 & 28.02 & \cellcolor{lightpurple}\textbf{28.40} \\
cat             & 26.18 & 25.21 & 26.18 & 24.48 & \cellcolor{lightpurple}\textbf{26.94}
                & 28.45 & 27.78 & 27.85 & \textbf{27.90} & \cellcolor{lightpurple}27.20 \\
cells\_battery  & 30.87 & 26.71 & 26.75 & 31.16 & \cellcolor{lightpurple}\textbf{31.58}
                & 29.61 & 29.46 & 28.72 & \textbf{30.94} & \cellcolor{lightpurple}30.15 \\
chair           & 30.51 & 20.55 & 20.57 & \textbf{29.66} & \cellcolor{lightpurple}29.31
                & 31.18 & 29.94 & 28.60 & \textbf{32.16} & \cellcolor{lightpurple}31.32 \\
chicken         & 28.31 & 21.44 & 21.78 & \textbf{27.47} & \cellcolor{lightpurple}27.43
                & 27.31 & 26.66 & 25.32 & \textbf{26.40} & \cellcolor{lightpurple}26.34 \\
cookie          & 29.98 & 27.08 & 29.99 & \textbf{31.74} & \cellcolor{lightpurple}30.58
                & 29.38 & 29.29 & 29.59 & 29.58 & \cellcolor{lightpurple}\textbf{30.23} \\
cupcake         & 29.17 & 20.63 & 21.68 & 23.69 & \cellcolor{lightpurple}\textbf{24.18}
                & 31.92 & \textbf{30.87} & 29.80 & 29.89 & \cellcolor{lightpurple}29.42 \\
deer\_moose     & 31.80 & 29.22 & 28.65 & 29.63 & \cellcolor{lightpurple}\textbf{30.26}
                & 30.29 & 28.92 & 28.30 & 29.45 & \cellcolor{lightpurple}\textbf{30.31} \\
donut           & 29.77 & 25.15 & 22.05 & \textbf{30.54} & \cellcolor{lightpurple}28.06
                & 32.71 & \textbf{31.72} & 31.63 & 31.29 & \cellcolor{lightpurple}31.15 \\
dragon          & 28.27 & 24.85 & 24.95 & 25.21 & \cellcolor{lightpurple}\textbf{25.24}
                & 26.22 & \textbf{26.68} & 25.11 & 26.26 & \cellcolor{lightpurple}25.97 \\
elephant        & 32.75 & 27.79 & 24.62 & \textbf{31.15} & \cellcolor{lightpurple}30.88
                & 32.57 & 30.35 & 29.36 & \textbf{31.32} & \cellcolor{lightpurple}31.08 \\
fish            & 28.71 & 25.62 & 25.56 & \textbf{27.66} & \cellcolor{lightpurple}27.39
                & 28.90 & 27.53 & 25.96 & \textbf{27.98} & \cellcolor{lightpurple}27.19 \\
flower          & 28.70 & 21.81 & 22.29 & 27.91 & \cellcolor{lightpurple}\textbf{28.09}
                & 25.46 & 25.25 & 25.12 & \textbf{25.59} & \cellcolor{lightpurple}25.37 \\
grapes          & 29.61 & 19.45 & 19.17 & 31.02 & \cellcolor{lightpurple}\textbf{31.07}
                & 25.24 & 24.36 & 24.55 & \textbf{25.45} & \cellcolor{lightpurple}25.07 \\
hammer          & 29.48 & 24.39 & 27.22 & \textbf{28.60} & \cellcolor{lightpurple}27.38
                & 27.27 & 28.86 & 28.70 & 28.70 & \cellcolor{lightpurple}\textbf{29.69} \\
helicopter      & 30.33 & 20.38 & 20.27 & 29.01 & \cellcolor{lightpurple}\textbf{30.03}
                & 25.10 & 23.34 & 22.56 & \textbf{24.22} & \cellcolor{lightpurple}22.52 \\
helmet          & 26.64 & 23.13 & 23.26 & 26.28 & \cellcolor{lightpurple}\textbf{27.11}
                & 24.42 & 23.87 & 23.78 & \textbf{24.06} & \cellcolor{lightpurple}23.96 \\
horse           & 28.13 & 23.33 & 23.28 & 25.38 & \cellcolor{lightpurple}\textbf{26.65}
                & 26.19 & 24.89 & 25.56 & 24.73 & \cellcolor{lightpurple}25.71 \\
ice\_cream      & 31.47 & 25.74 & 25.92 & 28.21 & \cellcolor{lightpurple}\textbf{30.07}
                & 34.74 & \textbf{35.20} & 35.17 & 33.46 & \cellcolor{lightpurple}34.27 \\
motorcycle      & 30.47 & 22.35 & 22.41 & \textbf{31.03} & \cellcolor{lightpurple}\textbf{31.03}
                & 27.06 & 24.91 & 24.70 & \textbf{25.47} & \cellcolor{lightpurple}25.33 \\
mouse           & 31.39 & 28.86 & 27.71 & \textbf{29.10} & \cellcolor{lightpurple}27.99
                & 29.56 & 27.92 & 28.45 & 28.45 & \cellcolor{lightpurple}\textbf{28.88} \\
mushroom        & 31.97 & 27.57 & 28.14 & 29.79 & \cellcolor{lightpurple}\textbf{29.89}
                & 34.40 & 33.92 & 32.14 & 33.93 & \cellcolor{lightpurple}\textbf{34.63} \\
orange          & 30.58 & 27.48 & 26.99 & 27.92 & \cellcolor{lightpurple}\textbf{28.03}
                & 29.54 & 30.37 & \textbf{30.63} & 29.48 & \cellcolor{lightpurple}29.37 \\
panda           & 29.64 & 27.27 & 26.64 & \textbf{31.62} & \cellcolor{lightpurple}30.94
                & 29.26 & 25.53 & 26.74 & 27.26 & \cellcolor{lightpurple}\textbf{28.72} \\
pear            & 33.55 & 28.20 & 28.44 & \textbf{33.33} & \cellcolor{lightpurple}32.99
                & 32.94 & 31.48 & 31.29 & 32.46 & \cellcolor{lightpurple}\textbf{33.67} \\
penguin         & 29.55 & 21.92 & 22.57 & \textbf{29.01} & \cellcolor{lightpurple}28.52
                & 29.44 & 27.00 & 27.15 & 29.38 & \cellcolor{lightpurple}\textbf{30.01} \\
piano           & 29.73 & 18.69 & 21.43 & \textbf{24.14} & \cellcolor{lightpurple}23.62
                & 16.98 & \textbf{19.86} & 19.19 & 19.77 & \cellcolor{lightpurple}18.24 \\
sandwich        & 29.23 & 22.64 & 25.30 & \textbf{27.53} & \cellcolor{lightpurple}24.81
                & 30.74 & 29.03 & 28.85 & \textbf{31.36} & \cellcolor{lightpurple}30.64 \\
screwdriver     & 28.08 & 26.28 & 25.62 & 26.08 & \cellcolor{lightpurple}\textbf{26.44}
                & 29.77 & \textbf{29.64} & 29.56 & 29.43 & \cellcolor{lightpurple}29.52 \\
shark           & 31.83 & 30.29 & 30.30 & 30.35 & \cellcolor{lightpurple}\textbf{30.65}
                & 31.92 & \textbf{31.33} & 31.17 & 30.91 & \cellcolor{lightpurple}31.02 \\
sheep           & 28.89 & 25.82 & 25.68 & 27.42 & \cellcolor{lightpurple}\textbf{28.43}
                & 28.07 & 28.46 & 27.99 & \textbf{29.13} & \cellcolor{lightpurple}28.86 \\
shoe            & 25.52 & 15.87 & 16.06 & \textbf{23.59} & \cellcolor{lightpurple}23.38
                & 22.41 & 21.13 & 19.90 & 22.70 & \cellcolor{lightpurple}\textbf{22.87} \\
snake           & 28.67 & 24.19 & 23.40 & \textbf{27.22} & \cellcolor{lightpurple}26.59
                & 27.61 & 24.00 & 23.80 & \textbf{25.27} & \cellcolor{lightpurple}25.03 \\
sofa            & 26.56 & 17.98 & 16.64 & 27.90 & \cellcolor{lightpurple}\textbf{27.94}
                & 26.46 & 25.51 & 24.55 & 25.45 & \cellcolor{lightpurple}\textbf{26.09} \\
tractor         & 31.25 & 27.66 & 26.10 & 26.82 & \cellcolor{lightpurple}\textbf{27.72}
                & 29.19 & 24.85 & 24.75 & \textbf{26.63} & \cellcolor{lightpurple}26.09 \\
train           & 22.65 & \textbf{26.15} & 25.00 & 22.18 & \cellcolor{lightpurple}22.36
                & 24.63 & \textbf{25.55} & 25.11 & 25.19 & \cellcolor{lightpurple}24.95 \\
trashcan        & 32.19 & 26.56 & 28.35 & 31.53 & \cellcolor{lightpurple}\textbf{32.14}
                & 30.26 & \textbf{32.05} & 31.46 & 31.31 & \cellcolor{lightpurple}31.70 \\
tree            & 27.98 & 18.59 & 18.98 & \textbf{28.13} & \cellcolor{lightpurple}27.94
                & 22.44 & 22.13 & \textbf{23.01} & 22.57 & \cellcolor{lightpurple}21.56 \\
violin          & 28.68 & 26.58 & 27.58 & \textbf{31.27} & \cellcolor{lightpurple}31.00
                & 28.51 & 27.73 & 27.32 & 27.65 & \cellcolor{lightpurple}\textbf{28.28} \\
\bottomrule
\end{tabular*}
\end{table*}

\begin{table*}[t]
\centering
\small
\caption{\textbf{Class-wise Evaluation on Novel-Class Assets (CLIP (↑))} for both TRELLIS-XL and Shap-E. Best scores are highlighted in \textbf{bold}, whereas our ReConText3D results are highlighted in purple.}
\label{tab:quant_classwise_novel_clip}
\setlength{\tabcolsep}{4pt}
\renewcommand{\arraystretch}{1.1}


\begin{tabular*}{\textwidth}{@{\extracolsep{\fill}}l cccc|cccc}
\toprule
\multirow{2}{*}{\textbf{Class}} &
\multicolumn{4}{c}{\textit{TRELLIS-XL}} &
\multicolumn{4}{c}{\textit{Shap-E}} \\
\cmidrule(lr){2-5} \cmidrule(lr){6-9}
& Fine-tuning & L2-SP & L2SP + Ours & \cellcolor{lightpurple}\textbf{Ours} &
Fine-tuning & L2-SP & L2SP + Ours & \cellcolor{lightpurple}\textbf{Ours} \\
\midrule

apple       & 33.14 & 34.09 & 32.03 & \cellcolor{lightpurple}\textbf{33.35}
            & 32.17 & 31.56 & 32.27 & \cellcolor{lightpurple}\textbf{32.37} \\
ball        & 29.32 & 31.29 & 31.24 & \cellcolor{lightpurple}\textbf{31.41}
            & 28.29 & 28.15 & 27.54 & \cellcolor{lightpurple}\textbf{28.44} \\
banana      & 33.03 & 33.00 & \textbf{33.15} & \cellcolor{lightpurple}33.03
            & 32.23 & 31.79 & 32.66 & \cellcolor{lightpurple}\textbf{33.14} \\
boat        & 29.36 & 27.53 & 29.00 & \cellcolor{lightpurple}\textbf{29.99}
            & \textbf{23.82} & 22.92 & 21.67 & \cellcolor{lightpurple}21.46 \\
bread       & 30.65 & \textbf{30.86} & 30.35 & \cellcolor{lightpurple}30.75
            & \textbf{27.42} & 26.68 & 26.10 & \cellcolor{lightpurple}26.01 \\
bunny       & 28.53 & \textbf{29.43} & 28.86 & \cellcolor{lightpurple}28.50
            & \textbf{29.94} & 29.26 & 28.67 & \cellcolor{lightpurple}29.30 \\
bus         & 26.24 & \textbf{28.38} & 27.16 & \cellcolor{lightpurple}27.57
            & \textbf{23.83} & 23.20 & 22.79 & \cellcolor{lightpurple}22.65 \\
butterfly   & \textbf{29.04} & 28.55 & 28.52 & \cellcolor{lightpurple}27.96
            & 27.03 & 27.00 & 27.77 & \cellcolor{lightpurple}\textbf{29.04} \\
chess\_piece & \textbf{36.07} & 35.81 & 34.40 & \cellcolor{lightpurple}35.82
            & 35.16 & 34.89 & \textbf{35.30} & \cellcolor{lightpurple}35.18 \\
coin        & 26.37 & \textbf{26.72} & 24.41 & \cellcolor{lightpurple}26.21
            & 25.80 & 24.73 & \textbf{26.72} & \cellcolor{lightpurple}25.29 \\
cow         & \textbf{32.96} & 32.43 & 30.41 & \cellcolor{lightpurple}31.40
            & 29.58 & \textbf{29.76} & 29.01 & \cellcolor{lightpurple}29.69 \\
crab        & 32.11 & 29.72 & \textbf{32.84} & \cellcolor{lightpurple}32.50
            & \textbf{28.92} & 27.78 & 27.25 & \cellcolor{lightpurple}28.57 \\
cup         & 29.57 & 29.93 & \textbf{29.98} & \cellcolor{lightpurple}29.60
            & 30.92 & 30.25 & \textbf{31.71} & \cellcolor{lightpurple}31.13 \\
dinosaur    & 27.63 & 27.23 & 26.59 & \cellcolor{lightpurple}\textbf{28.19}
            & \textbf{27.59} & 27.04 & 27.43 & \cellcolor{lightpurple}26.93 \\
dog         & \textbf{28.10} & 27.56 & 26.79 & \cellcolor{lightpurple}26.72
            & \textbf{27.21} & 27.20 & 27.06 & \cellcolor{lightpurple}27.13 \\
dolphin     & 32.09 & 31.94 & \textbf{32.42} & \cellcolor{lightpurple}31.79
            & 31.93 & 31.87 & 31.16 & \cellcolor{lightpurple}\textbf{32.11} \\
drum        & 26.58 & 26.52 & 26.68 & \cellcolor{lightpurple}\textbf{28.45}
            & 28.62 & 27.97 & \textbf{29.25} & \cellcolor{lightpurple}28.12 \\
fan         & \textbf{28.05} & 27.82 & 27.80 & \cellcolor{lightpurple}27.75
            & 25.12 & 24.77 & 25.46 & \cellcolor{lightpurple}\textbf{26.63} \\
fox         & 29.61 & \textbf{30.11} & 30.00 & \cellcolor{lightpurple}29.44
            & \textbf{29.58} & 28.68 & 29.52 & \cellcolor{lightpurple}28.88 \\
fridge      & \textbf{28.04} & 27.18 & 26.75 & \cellcolor{lightpurple}27.46
            & \textbf{25.63} & 25.39 & 25.22 & \cellcolor{lightpurple}25.21 \\
fries       & 30.96 & 30.81 & 31.41 & \cellcolor{lightpurple}\textbf{31.79}
            & 32.18 & 30.30 & 32.09 & \cellcolor{lightpurple}\textbf{32.41} \\
frog        & 30.05 & 29.98 & 30.89 & \cellcolor{lightpurple}\textbf{31.58}
            & \textbf{27.75} & 26.34 & 26.21 & \cellcolor{lightpurple}27.11 \\
glass       & 34.16 & 33.87 & \textbf{34.70} & \cellcolor{lightpurple}34.46
            & 32.90 & \textbf{33.08} & 32.38 & \cellcolor{lightpurple}32.97 \\
guitar      & 28.87 & 28.66 & \textbf{29.17} & \cellcolor{lightpurple}28.90
            & 28.25 & 28.17 & \textbf{28.33} & \cellcolor{lightpurple}28.08 \\
hamburger   & 32.83 & \textbf{33.30} & 31.95 & \cellcolor{lightpurple}33.12
            & \textbf{32.86} & 32.39 & 31.06 & \cellcolor{lightpurple}30.97 \\
hat         & \textbf{28.28} & 25.74 & 27.51 & \cellcolor{lightpurple}27.31
            & 27.28 & 26.63 & 27.69 & \cellcolor{lightpurple}\textbf{27.72} \\
key         & 28.37 & \textbf{28.94} & 28.12 & \cellcolor{lightpurple}28.50
            & 28.89 & 28.84 & 29.90 & \cellcolor{lightpurple}\textbf{30.01} \\
knife       & 27.53 & 27.23 & \textbf{27.80} & \cellcolor{lightpurple}26.77
            & 28.33 & 27.89 & 27.96 & \cellcolor{lightpurple}\textbf{28.33} \\
laptop      & 29.51 & 29.92 & 30.11 & \cellcolor{lightpurple}\textbf{30.15}
            & 26.11 & 26.01 & 26.84 & \cellcolor{lightpurple}\textbf{26.91} \\
lizard      & \textbf{27.48} & 27.00 & 27.43 & \cellcolor{lightpurple}26.57
            & \textbf{26.45} & 25.97 & 25.87 & \cellcolor{lightpurple}25.47 \\
monkey      & 29.55 & \textbf{29.62} & 28.48 & \cellcolor{lightpurple}28.92
            & 26.75 & 26.75 & \textbf{27.00} & \cellcolor{lightpurple}26.88 \\
mug         & 30.15 & 29.81 & 30.09 & \cellcolor{lightpurple}\textbf{30.24}
            & \textbf{31.34} & 31.12 & 31.01 & \cellcolor{lightpurple}31.23 \\
octopus     & \textbf{31.38} & 29.15 & 29.30 & \cellcolor{lightpurple}30.19
            & 28.12 & 28.18 & \textbf{28.96} & \cellcolor{lightpurple}28.68 \\
pencil      & 31.27 & 31.81 & 30.87 & \cellcolor{lightpurple}\textbf{32.67}
            & 30.45 & 30.00 & 31.16 & \cellcolor{lightpurple}\textbf{31.31} \\
phone       & 26.88 & 21.89 & 24.50 & \cellcolor{lightpurple}\textbf{27.33}
            & 21.47 & 22.70 & 21.14 & \cellcolor{lightpurple}\textbf{22.07} \\
pig         & \textbf{33.83} & 32.20 & 33.22 & \cellcolor{lightpurple}32.19
            & \textbf{32.97} & 32.32 & 32.18 & \cellcolor{lightpurple}31.89 \\
pizza       & 31.12 & \textbf{31.87} & 31.32 & \cellcolor{lightpurple}29.56
            & \textbf{32.77} & 32.44 & 32.43 & \cellcolor{lightpurple}32.41 \\
plate       & 31.57 & 32.08 & \textbf{33.12} & \cellcolor{lightpurple}32.46
            & 32.55 & 32.08 & \textbf{33.13} & \cellcolor{lightpurple}33.06 \\
radio       & 24.07 & \textbf{25.56} & 23.82 & \cellcolor{lightpurple}25.45
            & 24.71 & 23.39 & \textbf{24.99} & \cellcolor{lightpurple}23.82 \\
robot       & 29.16 & 28.70 & 28.14 & \cellcolor{lightpurple}\textbf{30.08}
            & \textbf{29.02} & 28.49 & 26.81 & \cellcolor{lightpurple}27.47 \\
sink        & \textbf{31.12} & 31.11 & 30.05 & \cellcolor{lightpurple}31.11
            & 27.95 & 27.01 & \textbf{29.23} & \cellcolor{lightpurple}28.43 \\
spade       & 28.88 & 29.72 & 29.12 & \cellcolor{lightpurple}\textbf{30.08}
            & \textbf{30.71} & 30.08 & 29.21 & \cellcolor{lightpurple}30.07 \\
stove       & 28.69 & \textbf{29.70} & 28.24 & \cellcolor{lightpurple}28.80
            & 25.34 & 25.16 & 25.37 & \cellcolor{lightpurple}\textbf{25.62} \\
truck       & 27.75 & 26.40 & \textbf{28.08} & \cellcolor{lightpurple}25.59
            & \textbf{24.89} & 24.76 & 24.65 & \cellcolor{lightpurple}24.01 \\
whale       & 31.33 & 32.01 & 31.42 & \cellcolor{lightpurple}\textbf{32.16}
            & 30.47 & 31.23 & \textbf{31.69} & \cellcolor{lightpurple}31.01 \\
\bottomrule

\end{tabular*}
\end{table*}

\section{Additional Qualitative Results}
\label{section:supp_qualitative_results}
\begin{figure*}
  \centering
    \includegraphics[width=1\linewidth]{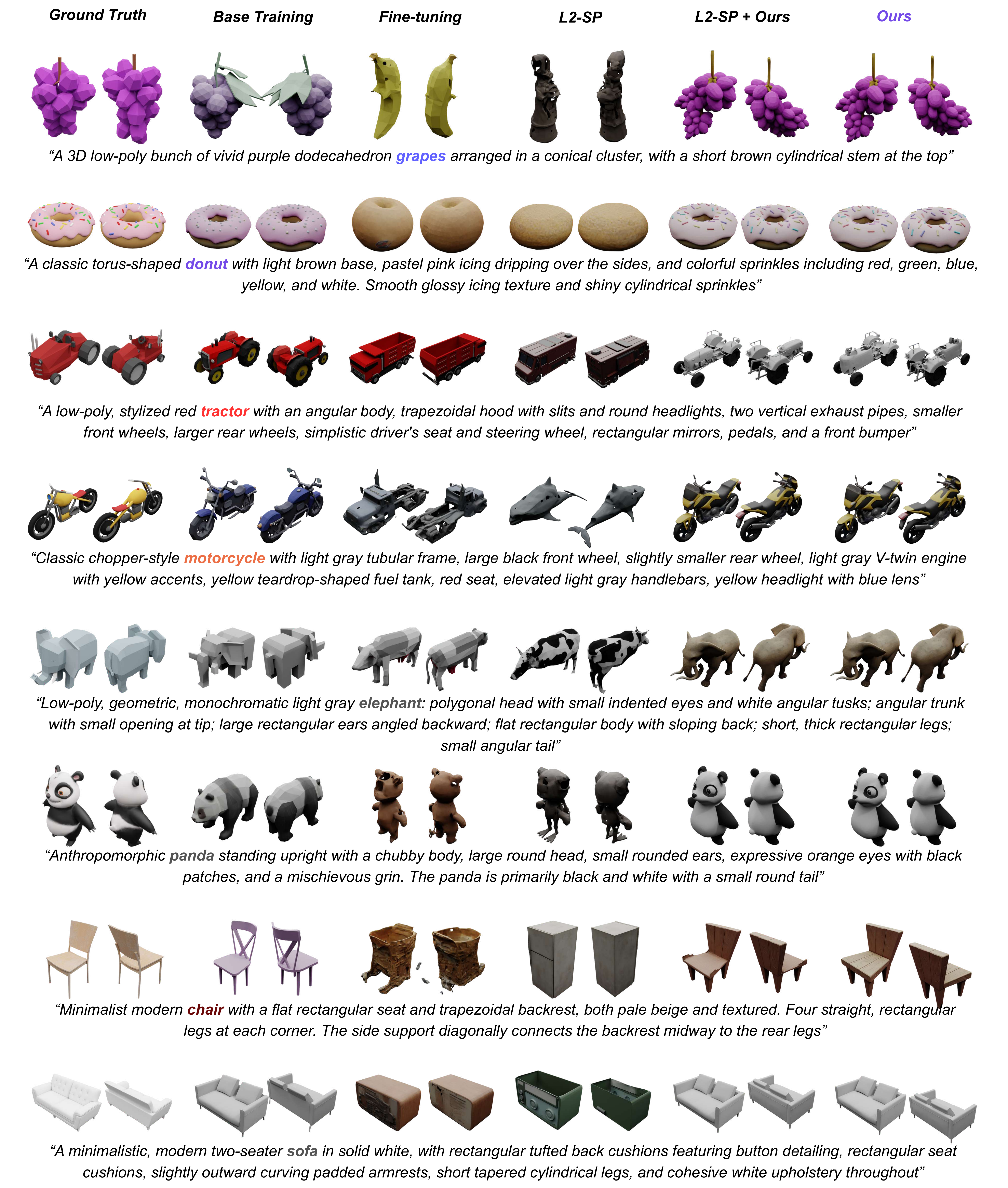}
    \caption{Qualitative comparison of continual text-to-3D baselines against ReConText3D (Ours) on \textbf{base class} assets using \textbf{TRELLIS-XL}.}
    \label{fig:sup_trellis_stage1}
   
\end{figure*}

\begin{figure*}
  \centering
    \includegraphics[width=1\linewidth]{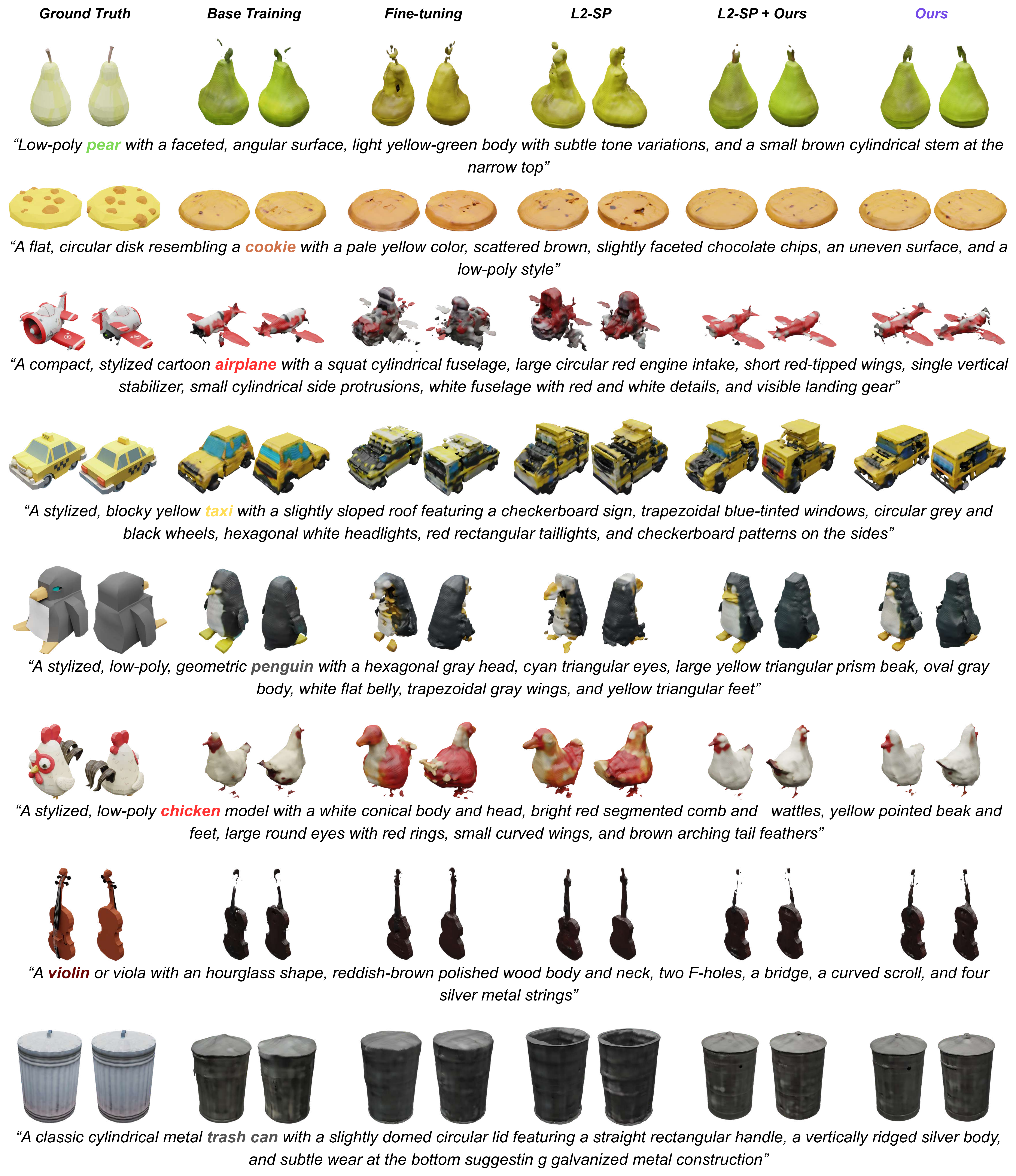}
    \caption{Qualitative comparison of continual text-to-3D baselines against ReConText3D (Ours) on \textbf{base class} assets using \textbf{Shap-E}.}
    \label{fig:sup_shape_stage1}
   
\end{figure*}

In this section, we provide extended qualitative comparisons. While CLIP scores are a useful metric for semantic alignment, qualitative inspection remains essential for assessing geometric fidelity, texture realism, structural consistency, and the extent of catastrophic forgetting in continual text-to-3D generation. 
\paragraph{Base-class performance.}
Figures~\ref{fig:sup_trellis_stage1} and~\ref{fig:sup_shape_stage1} visualize generations for base classes on TRELLIS-XL and Shap-E, respectively. When fine-tuned on novel classes, both models exhibit clear signs of catastrophic forgetting, but in distinct ways. TRELLIS-XL frequently mixes or shifts the semantic class of the object, indicating that category-level information is easily overwritten during continual updates. For example, in row~2 of Figure~\ref{fig:sup_trellis_stage1}, the fine-tuned TRELLIS-XL model, and even L2-SP, changes a \emph{donut} into a \emph{bread}-like object. In row~3 both these models again alter a \emph{tractor} into something closer to a \emph{truck}. In contrast, our ReConText3D approach correctly preserves the original base-class semantics in both cases, reliably generating a \emph{donut} and a \emph{tractor} with appropriate geometry and structure even after novel-class training.

Shap-E behaves slightly differently. It tends to retain the correct class identity more often but suffers from degraded geometry, yielding structurally weakened outputs. Nonetheless, Shap-E is not fully immune to semantic drift. Row~3 of Figure~\ref{fig:sup_shape_stage1} shows a notable case where the model appears to confuse an \emph{airplane} with a \emph{tractor}, blending features from both categories.

\begin{figure*}
  \centering
    \includegraphics[width=1\linewidth]{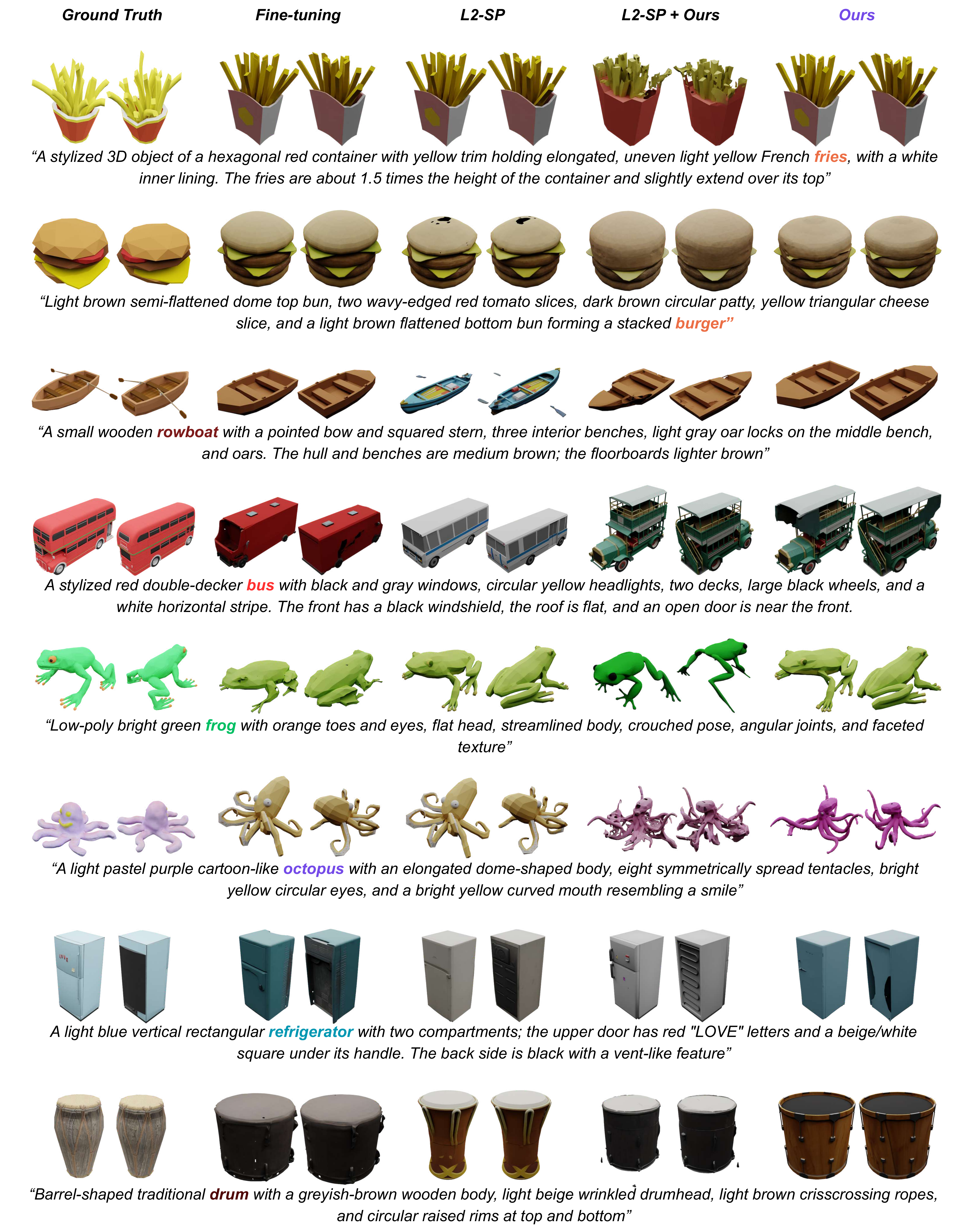}
    \caption{Qualitative comparison of continual text-to-3D baselines against ReConText3D (Ours) on \textbf{novel class} assets using \textbf{TRELLIS-XL}. }
    \label{fig:sup_trellis_stage2}
   
\end{figure*}

\begin{figure*}
  \centering
    \includegraphics[width=1\linewidth]{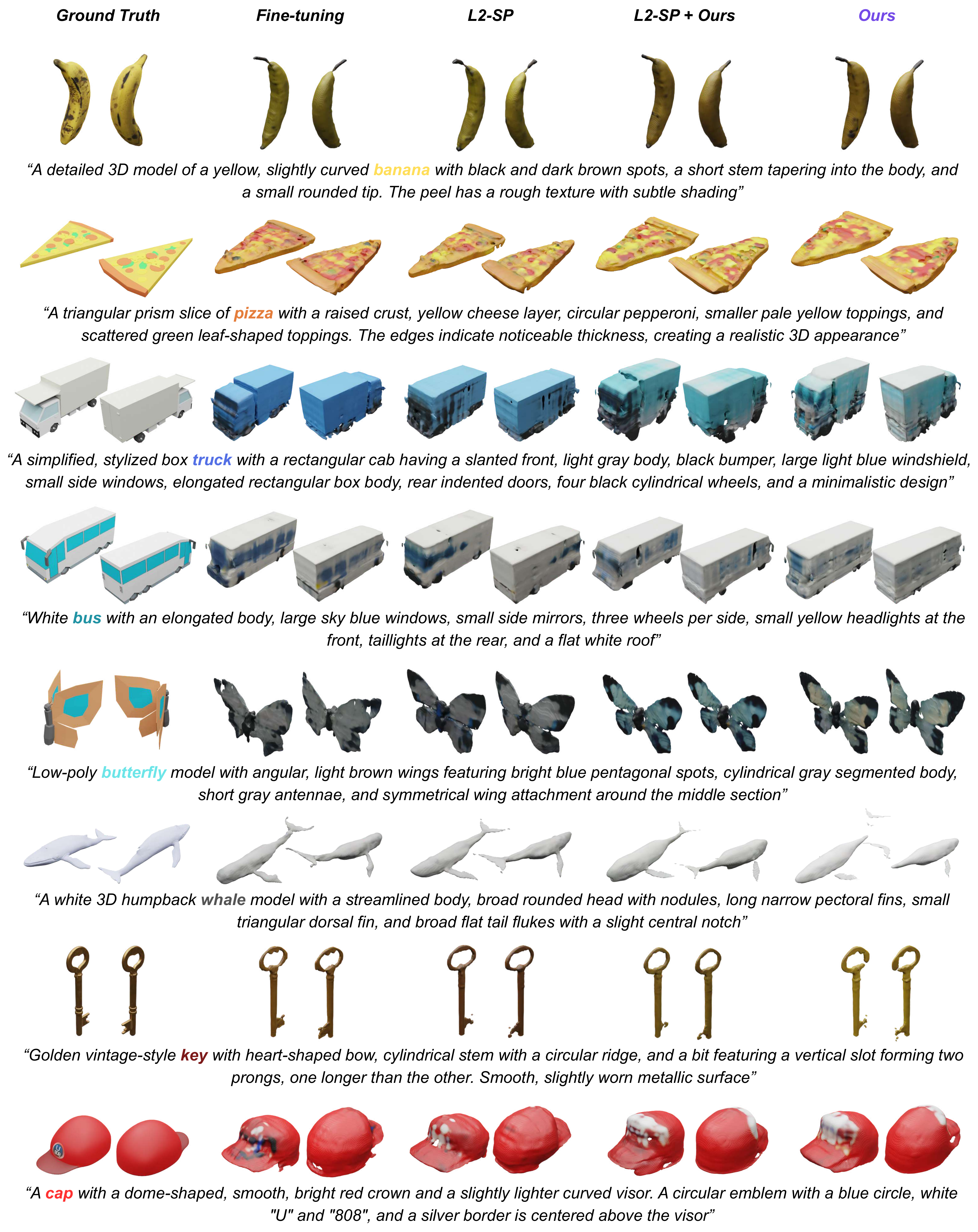}
    \caption{Qualitative comparison of continual text-to-3D baselines against ReConText3D (Ours) on \textbf{novel class} assets using \textbf{Shap-E}.}
    \label{fig:sup_shape_stage2}
   
\end{figure*}

\paragraph{Novel-class performance.}
Figures~\ref{fig:sup_trellis_stage2} and~\ref{fig:sup_shape_stage2} present qualitative results for novel classes on TRELLIS-XL and Shap-E. Overall, fine-tuning performs well on many newly introduced categories and often produces good stage-2 assets. However, there are notable cases where our ReConText3D approach surpasses the fine-tuned models in visual fidelity or semantic accuracy. For example, in row~6 of Figure~\ref{fig:sup_trellis_stage2}, the fine-tuned TRELLIS-XL model generates an \emph{octopus} with an incorrect orange coloration, whereas our method produces the intended purple version with a more coherent overall structure. 


\paragraph{Failure Cases.}

While ReConText3D largely preserves semantic and structural fidelity for both base and novel classes, there are still occasional failure modes. For instance, Figure~\ref{fig:sup_trellis_stage1} row~3 shows a \emph{tractor} that is correctly classified but generated in grey rather than the intended red, indicating a loss of color information. Similarly, Figure~\ref{fig:sup_trellis_stage2} row~4 depicts a \emph{bus} generated in green instead of red.

These examples highlight that while our semantic replay strategy effectively stabilizes class identity and structure, it can sometimes fail to retain fine-grained visual attributes such as color. Addressing such limitations could further enhance the fidelity of stage-1 and stage-2 assets in future work.

\end{document}